\documentclass{article}
\usepackage{arxiv}
\usepackage[utf8]{inputenc} 
\usepackage[T1]{fontenc}    
\usepackage{hyperref}       
\usepackage{url}            
\usepackage{booktabs}       
\usepackage{amsfonts}       
\usepackage{nicefrac}       
\usepackage{microtype}      
\usepackage{lipsum}		
\usepackage{graphicx}
\usepackage[square,numbers]{natbib}
\usepackage{doi}
\usepackage{colortbl}
\usepackage{algorithm}
\usepackage{algorithmicx}
\usepackage{algpseudocode}
\usepackage{amsmath}
\usepackage[normalem]{ulem}
\usepackage{amsfonts}
\usepackage{tabu}                      
\usepackage{booktabs}
\usepackage{array}
\definecolor{inchworm}{rgb}{0.7, 0.93, 0.36}

\newcolumntype{P}[1]{>{\centering\arraybackslash}p{#1}}
\newcolumntype{M}[1]{>{\centering\arraybackslash}m{#1}}

\title{Prediction of Dilatory Behavior in eLearning: A
Comparison of Multiple Machine Learning Models}

\author{Christof Imhof,
        Ioan-Sorin Comsa,
        Martin Hlosta,
        Behnam Parsaeifard,
        Ivan Moser, and
        Per Bergamin}

\renewcommand{\shorttitle}{Prediction of Dilatory Behavior in eLearning}

\hypersetup{
pdftitle={Prediction of Dilatory Behavior in eLearning: A
Comparison of Multiple Machine Learning Models},
pdfsubject={q-bio.NC, q-bio.QM}
}

\begin{document}
\maketitle

\begin{abstract}
Procrastination, the irrational delay of tasks, is a common occurrence in online learning. Potential negative consequences include higher risk of drop-outs, increased stress, and reduced mood. Due to the rise of learning management systems and learning analytics, indicators of such behavior can be detected, enabling predictions of future procrastination and other dilatory behavior. However, research focusing on such predictions is scarce. Moreover, studies involving different types of predictors and comparisons between the predictive performance of various methods are virtually non-existent. In this study, we aim to fill these research gaps by analyzing the performance of multiple machine learning algorithms when predicting the delayed or timely submission of online assignments in a higher education setting with two categories of predictors: subjective, questionnaire-based variables and objective, log-data based indicators extracted from a learning management system. The results show that models with objective predictors consistently outperform models with subjective predictors, and a combination of both variable types perform slightly better. For each of these three options, a different approach prevailed (Gradient Boosting Machines for the subjective, Bayesian multilevel models for the objective, and Random Forest for the combined predictors). We conclude that careful attention should be paid to the selection of predictors and algorithms before implementing such models in learning management systems.
\end{abstract}


C. Imhof, I.-S. Comsa, M. Hlosta, B. Parsaeifard, I. Moser, and P. Bergamin are with the Institute for Research in Open-, Distance- and eLearning, Swiss Distance University of Applied Sciences, Brig, CH-3900, Switzerland (e-mail addresses: christof.imhof@ffhs.ch, ioan-sorin.comsa@ffhs.ch, martin.hlosta@ffhs.ch, behnam.parsaeifard@ffhs.ch, ivan.moser@ffhs.ch, per.bergamin@ffhs.ch). P. Bergamin is also affiliated with the North-West University, Potchefstroom, 2531, South Africa.

\section{Introduction}
The ubiquity of digitization continues to pervade many aspects of life, including learning. While this ongoing trend entails numerous advantages for students, such as more flexible schedules, more engaging distance education or a greater variety of interactive learning tasks \cite{Dumford2018}, drawbacks also need to be considered. One potential drawback is the increased likelihood of academic procrastination due to the many distractions our digital world offers. \emph{Academic procrastination} is a type of dilatory behavior that is commonly defined as the irrational delay of academic tasks such as writing assignments or studying course literature \cite{Steel2007}. While seemingly harmless, procrastination may lead to various negative consequences, which include increased stress, lower performance, and reduced mood \cite{Tice1997}. Moreover, this behavior is linked to increased risk of drop-out \cite{doherty2006} and is reportedly highly prevalent among students in higher education \cite{ellis1977,Steel2007}. 

However, not all acts of delay are necessarily maladaptive since they can also be used as a deliberate strategy. The positive, productive counterpart to procrastination is known as \emph{purposeful delay} \cite{corkin2011comparing}. Alternative names for this concept include active delay, strategic delay or active procrastination, the latter of which is considered to be an oxymoron \cite{Pychyl2008}. The term \emph{dilatory behavior} thus serves as an umbrella term that refers to both positive and negative forms of delay.

A promising approach to measuring indicators of dilatory behavior comes in the form of \emph{Learning Analytics} (LA). LA is often implemented as a conceptual framework to analyse course characteristics, such as prediction of students’ learning performance, educational data analysis, data collection and measurement, and early intervention \cite{la_hwang_2017}. Given the potential consequences of negative dilatory behavior, the ability to detect it early utilizing an LA framework could prove very useful to teachers, lecturers, and students alike. This especially concerns learning management systems (LMS), which are able to provide objective indicators for such behavior. The earlier the detection, the more time remains to intervene and to ultimately prevent potential drop-outs.

Predicting dilatory behavior in academic tasks is not trivial for multiple reasons, especially in the field: 
various courses must be incorporated to allow for a generalization of such predictions and their timing also needs to be considered. In this context, data collection and the type of variables (be they predictors or outcome variables) play a crucial role. 
Despite the variety of types of \emph{Machine Learning} (ML) models that could be employed, their predictive performance may vary as a result of the size of the collected data set, the number of predictor variables, and the point in time predictions are made. Therefore, the selection of an appropriate prediction model must involve such considerations.

Previous studies indicate the great potential of ML-based prediction models in learning analytics \cite{{dt_azimi_2020},{pca_pardo_2016}, {olsr_conijn_2016}}, even when predicting dilatory behavior \cite{Akram2019}. To our knowledge, however, comparisons between multiple ML algorithms are rather rare in this context (see \cite{Akram2019} and \cite{Cerezo2017}). Moreover, delay is usually treated as a predictor for other outcome variables rather than being the outcome variable itself. To fill this gap, we employ eight different ML algorithms to predict dilatory behavior in online assignments. For our analyses, we used the same data set as in a prior study, where the prediction of dilatory behavior was investigated, comparing models with subjective and objective predictors using Bayesian multilevel regression \cite{Imhof2021}. We provide a comparative analysis to determine which algorithms deliver the best predictions and compare the performance with the Bayesian models as a baseline.

\section{Related Work}
In this section, we present the related work on dilatory behavior and the machine learning algorithms employed to predict such tendencies in various tasks. The analytical framework, research questions, and contributions of this paper are also detailed. 

\vspace{-0.05in}
\subsection{Theoretical Background of Dilatory Behavior}
As stated in the introduction, dilatory behavior is an umbrella term that encompasses at least two types of delay, procrastination and purposeful delay. However, another important distinction should be made when investigating delay, namely the one between trait and state delay. \emph{State} variables are situational, whereas \emph{trait} variables are stable with less situation- or time-specific variance. In the case of procrastination, the correlation between trait and state (or in other words, attitude and actual behavior) is reportedly medium to large at .51 \cite{Steel2007}. This not only implies that situational factors are important when trying to predict delay, but also that trait procrastination can still serve as a predictor for state procrastination (or vice versa), albeit not a particularly strong one.

This train of thought leads to the question, which additional predictors besides trait procrastination should then be considered when predicting procrastination and other types of dilatory behavior. Two broad categories present themselves: subjective and objective predictors. The former are commonly assessed with questionnaires, while the latter can be gained by observation, or in the case of LMS, collection of log data. 

When predicting delay with subjective predictors, motivational aspects and learning-related factors suggest themselves. One option are factors in the orbit of self-regulation, given that procrastination is often being considered a "self-regulation failure" \cite{Steel2007} whereas purposeful delay appears as its successful counterpart \cite{Corkin2011}. Learning concepts related to self-regulation include academic self-efficacy and self-directed learning. \emph{Academic self-efficacy} refers to the belief in one's ability to succeed at academic tasks such as writing exams \cite{Bandura1986} and is negatively linked to procrastination \cite{Waschle2014a}, forming a vicious circle, but positively connected to purposeful delay, counteracting that circle \cite{Chu2005}.

\emph{Self-directed learning} (SDL) is a process in which students take responsibility for their own learning, which involves self-monitoring, self-management, and self-evaluation of the learning process \cite{Bolhuis1996, Knowles1975, Garrison1997}. Research reveals positive links between SDL and self-efficacy \cite{Saeid2017} and negative links with procrastination \cite{Schommer2018cognitive}.

In the context of LMS, it is self-evident to consider objective predictors based on data extracted from such systems. Popular approaches in learning analytics are to analyze click-based data (e.g., \cite{Cirigliano2020, Knight2017}), which provide insights into student activity and engagement with learning content on the platform.

\vspace{-0.05in}
\subsection{Delay-related Prediction Models in Learning Analytics}
Prediction models that involve delay as a predictor are quite common in the literature. A variety of methods have been implemented to investigate the relationship between procrastination (as the most commonly researched type of delay) and other variables, usually achievement. A first example is \cite{You2015}, who used multiple regression analysis to investigate the effects of procrastination on course achievement. Their regression model predicted achievement at each point in time, with the predictability increasing as time passed. The authors in \cite{Levy2012} instead implemented data analytics techniques to detect anomalies in their data, corroborating the finding that procrastination negatively predicts performance (grades). The same conclusion was reached by the authors in \cite{Gareau2019}, who used Structure Equation Modeling and discovered a mediating role of task-oriented and disengagement-oriented coping. Another technique, path analysis, was applied in \cite{Yamada2016}, where the authors found that positive time management promoted early submissions of paper reports. A different method, association rule mining, was implemented in \cite{DelPuertoPaule-Ruiz2015} and \cite{Cerezo2017}. Both research groups identified indicators of procrastination on Moodle, which were used to create association rules in the form of conditionals by implementing either Apriori and/or Predictive Apriori algorithms. In both studies, time-related indicators were closely related to students' performance. 
In contrast, models that explicitly try to predict delay as the outcome variable are rather scarce. Exceptions include studies conducted in \cite{Akram2019}, \cite{Hooshyar2020mining}, \cite{abidi2020educational}, and \cite{Yang2020a}, who all intended to classify students based on homework submission data. Ten different ML algorithms (ZeroR, OneR, ID3, J48, Random Forest, decision stump, JRip, PART, NBTree, and Prism) were implemented in \cite{Akram2019} to classify students as procrastinators or non-procrastinators based on feature vectors. The optimal amount of clusters was determined to be three (one non-procrastinating group and two procrastinating classes that differ in submission scores).

Multiple clustering algorithms were also used in \cite{Hooshyar2020mining} and \cite{Yang2020a} to detect the optimal amount of clusters, followed by a classification of students into the three resulting clusters (procrastinators, non-procrastinators, and procrastinator candidates). In both studies, the authors compared eight methods of classification (linear and radial basis function kernel support vector machine, Gaussian Processes, decision tree, Random Forest, neural network, AdaBoost, and Naive Bayes). The former implemented a feature vector algorithm involving categorical and continuous features based on spare time (i.e., interval between submission and deadline) and inactive time (i.e., time before the first click on an assignment), and found that neural networks worked best with categorical features and that linear support vector machines outperformed the others in the case of continuous features. The latter reported that the linear support vector machine approach delivered the best predictive performance for their clusters.

Binary classifiers were employed in \cite{abidi2020educational}, using four supervised-learning algorithms (logistic regression, decision tree, gradient boosting, and Random Forest) to classify students as procrastinators or non-procrastinators based on data that was extracted from an intelligent tutoring system (ITS). Among the ML algorithms, gradient boosting had the best performance in terms of classification precision.
However, these examples involve a classification of students into clusters, rather than predictions regarding the extent of the delay itself. This research gap was one of the reasons we conducted a previous study to determine if dilatory behavior could be predicted based on a mixture of questionnaire scores and log data \cite{Imhof2021}.

In that study, seven predictors were implemented across six different Bayesian multilevel models to determine which type of predictor (objective vs. subjective) and which individual predictors were able to predict dilatory behavior the best. The four subjective predictors were questionnaire scores and the three objective predictors were based on log data. The model fit comparison favored the models that included all seven predictors, but their advantage over the models that only included objective predictors was minimal at best. This implies that the models with objective predictors barely improve when subjective predictors are added. However, this result does not imply that subjective predictors should be discarded altogether. First off, the results may not be the same when following other approaches for prediction models, for instance models based on ML. Secondly, we focused on individual predictors and comparisons between types of predictors rather than analyzing the actual performance of the predictions in terms of accuracy and other measures. 

\vspace{-0.05in}
\subsection{Research Questions}
Therefore, the goal of this study was to extend the findings reported in \cite{Imhof2021} by comparing and contrasting the performance of different approaches to prediction models, determining which ML algorithm delivers the best predictions for delay and for which type of predictor.

Before the predictive performance of these models can be assessed, it needs to be clarified how the hyperparameters for each ML algorithm and type of predictor must be determined in order for them to be optimized. We present a novel cross-validation approach to address this and compare the results of the cross-validation with the rest results. We then intend to identify the ML model with the best predictive performance for the following three predictor types: subjective (\texttt{subj}), objective (\texttt{obj}), and a combination between the two (\texttt{comb}). Finally, we determine whether models with objective predictors still outperform models with subjective predictors when calculated with various ML algorithms and whether there is still an advantage of models that combine both types of predictors for all of our algorithms. The research questions are thus as follows:

\textbf{RQ1:} How well do the results of the cross-validation with the proposed measure compare to the test results?

\textbf{RQ2:} Which machine learning algorithm delivers an improvement of the predictive performance compared to the baseline models based on the subjective, objective, and combined sets of predictors (intra-model comparison)?

\textbf{RQ3:} Which type of predictors (subjective, objective, and combined) allow for the highest predictive performance (inter-model comparison)? 

\subsection{Paper Organization}
The rest of the paper is organized as follows: In Section III, we present the methods and instruments used to collect the data. Then, we introduce the prediction framework and briefly illustrate each ML algorithm. A novel cross-validation procedure is also presented to determine the best set of hyperparameters for all of the models. Next, we present our results in Section IV, which includes the results of the cross-validations and the test results of each ML model, which are compared with the performance of the Bayesian multilevel models. This is followed by a discussion of our results in Section V, including implications and limitations of our findings. Finally, the paper concludes with Section VI.

\section{Methods and Prediction Models}
\subsection{Participants and Courses}
We used the same data set as in a prior study, which included 134 students from a distance university located in Central Europe \cite{Imhof2021}. The sample included 65 male and 69 female participants (mean age 31.61 years, sd = 7.97, min = 19, max = 61). The students were enrolled in at least one course each during the autumn semester of 2019. The number of involved online courses was 126, each belonging to one of three departments (computer science, economics, and health). Every course consisted of several blocks (usually between five and ten), each with their own assignment(s). The total number of assignments across all students and courses was 1107. The students participated voluntarily and consented to have their log data extracted from the institution's LMS by filling in an online survey. As compensation, all participants automatically entered a raffle with a chance of winning cinema vouchers. 

\vspace{-0.1in}
\subsection{Instruments, Procedures, and Variables}
As described in \cite{Imhof2021}, the procedure started with an e-mail invitation that was sent to all students enrolled at our institution at the end of the semester. The volunteers then followed a link to an online survey, consisting of four questionnaires, whose scores formed the subjective predictors for our models: the General Academic Self-Efficacy (GASE) Scale \cite{Nielsen2018}, consisting of five items, the Self-Directed Learning Scale (SDLS) \cite{Lounsbury2009} with ten items, the Academic Procrastination Scale - short form (APSS) \cite{McCloskey2011} with five items, and the Active Procrastination Scale (APS) \cite{Choi2009} with four subscales and a total of 16 items.

\begin{table*}[t]
    \caption{Description of Subjective and Objective Predictors and Outcome variable}
    \label{tab:my_label}
    \centering
    \begin{tabular}{|c|l|l|l|l|l|l|}
         \hline
         \multicolumn{1}{m{3cm}|}{Variable} & Role & Mean & SD & Min & Max & Description \\
         \hline
         \multicolumn{1}{m{3cm}|}{GASE} & \multicolumn{1}{m{3cm}|} {subjective predictor} & 19.71 & 3.10 & 8 & 25 & \multicolumn{1}{m{4cm}|}{General Academic Self-Efficacy Scale, measures belief in one's academic abilities} \\ 
         \hline
         \multicolumn{1}{m{3cm}|}{SDLS} & \multicolumn{1}{m{3cm}|} {subjective predictor} & 38.74 & 5.29 & 18 & 50 & \multicolumn{1}{m{4cm}|}{Self-Directed Learning Scale, measures how much one feels in charge of one's learning process} \\
         \hline
         \multicolumn{1}{m{3cm}|}{APSS} & \multicolumn{1}{m{3cm}|} {subjective predictor} & 11.04 & 4.16 & 5 & 21 & \multicolumn{1}{m{4cm}|}{Academic Procrastination Scale – Short Form, measures tendency to procrastinate in academic tasks} \\
         \hline
         \multicolumn{1}{m{3cm}|}{APS} & \multicolumn{1}{m{3cm}|} {subjective predictor} & 72.51 & 11.87 & 43 & 104 & \multicolumn{1}{m{4cm}|}{Active Procrastination Scale, measures purposeful delay, i.e., tendency to delay tasks strategically}\\
         \hline
         \multicolumn{1}{m{3cm}|}{Number of clicks on assignment} & \multicolumn{1}{m{3cm}|} {objective predictor} & 6.58 & 4.91 & 1 & 34 & \multicolumn{1}{m{4cm}|}{Sum of clicks on an assignment before submission} \\
         \hline
         \multicolumn{1}{m{3cm}|}{Interval between start of block and first click on assignment (days)} & \multicolumn{1}{m{3cm}|}{objective predictor} & -7.13 & 32.40 & -150.30 & 98.72 & \multicolumn{1}{m{4cm}|}{Difference between start of a block (learning unit on Moodle) and the first click on an assignment (days)}\\
         \hline
         \multicolumn{1}{m{3cm}|}{Number of clicks on relevant activities in the course} & \multicolumn{1}{m{3cm}|} {objective predictor} & 173.83 & 168.44 & 0 & 1237 & \multicolumn{1}{m{4cm}|}{Sum of all clicks on activities (quizzes, videos, books, etc.), indicates general activity in the course}\\
         \hline
         \multicolumn{1}{m{3cm}|}{Delay (days)} & outcome variable & -1.66 & 18.26 & -113.51 & 132.43 & \multicolumn{1}{m{4cm}|}{Difference between deadline of an assignment and time of submission, positive values indicating delay and negative values meaning early submissions}\\
         \hline
    \end{tabular}
\end{table*}
 
The three objective predictors, which the authors selected based on log data variables used in other studies \cite{DelPuertoPaule-Ruiz2015, Cerezo2017}, were the number of clicks on an assignment, the interval between the start of a block and the first click on an assignment, and the number of clicks on relevant activities in the course. The \emph{number of clicks on an assignment} reflects the sum of all clicks on the assignment section of the course made before the deadline had passed. The second predictor, the \emph{interval between the start of a block and the first click on an assignment}, indicates the time (in days) that passed between when a block had started and when the first click on the assignment was made (i.e., when the task description was first read). The final objective predictor was the \emph{number of clicks on relevant activities in the course}, which reflected the overall engagement with the course material on the platform (i.e., the sum of all clicks on learning videos, forums, interactive books, etc.). The interval turned out to be the strongest and most consistent individual predictor in the previous study, followed by the number of clicks on the assignment.

The to-be-predicted outcome variable was \emph{delay}, which can assume positive and negative values, with positive values indicating a delayed submission of an assignment (in hours) and negative values meaning a timely submission. In this study, \emph{delay} was used for regression and classification alike, the former to determine the error between predicted and actual values, and the latter to determine whether the two classes (delay vs. timely submission) would be correctly predicted.

In Table I, we provide an overview of the characteristics of the outcome variable, the four subjective, and the three objective predictors (i.e., their role in the models, their mean, standard deviation, minimum and maximum values, and a brief description). These characteristics are very useful in adopting different normalization techniques (e.g., standard score or min-max feature scaling). In this paper, we adopt the max-absolute normalization technique to have the block-click-interval and delay features in the interval of [-1, 1].

\vspace{-0.05in}
\subsection{Machine Learning Models}
Before training the models, we first removed all missing values (e.g., assignments that were not handed in at all) and then normalized the  data. The processed data set (1107 rows, each representing one individual assignment) was then split into training and testing sets, the former including 80\% of the data (885 rows) and the latter 20\% (222 rows). Since the original data set is rather small, the subsets could differ in their distributions depending on the way the data happens to be split. We addressed this concern by repeating the randomized splitting process ten times, thus creating ten pairs of training and testing sets. When individually compared, the testing subsets shared 44.7 rows on average (sd = 5.69, min = 31, max = 57), ruling out potential split-related biases. For further processing, we then split all training and testing sets into three subsets each based on the type of predictor: a subset with subjective predictors, one with objective predictors, and a subset with both types of predictors combined. All sets and subsets were of equal length, the only difference being the included predictors.

For every training set, we then determined the hyperparameters of each ML algorithm based on cross-validation. Once the models' hyperparameters were fixed, we trained each ML algorithm by individually exposing them to all ten training data sets to learn predicting the outcome variable \emph{delay} based on the subjective, objective or both sets of predictors combined. Our objective was twofold: predicting real values in terms of delay and assessing how accurate the classifications were. Therefore, the ML algorithms to be analyzed in this paper all needed to be able to produce both regression and classification models. We thus chose the following eight algorithms: Naive Bayes (NB), K-Nearest neighbors (KNN), Radial Basis Function Networks (RBFN), Feed-Forward Neural Networks (FFNN), Regression Trees (RT), Gradient Boosting Machines (GBM), Random Forests (RF), and Support Vector Regression (SVR). Out of this selection, we aimed to determine the algorithms that \textit{a)} minimized the error between the predicted and actual delay values from the testing data sets; and \textit{b)} obtained the best classification performance when predicting the delay and timely-submission classes. In the following, we introduce each of these techniques, starting with the Bayesian multilevel models that serve as the baseline for our comparisons.

\subsubsection{Bayesian Multilevel Models}
We selected Bayesian multilevel regression models as the baseline for our comparisons since they were already implemented in \cite{Imhof2021}, whose data this study is based on. In that study, Bayesian multilevel models were chosen to match the nested data structure (students being enrolled in multiple courses, each with their own assignments). The authors favored a Bayesian approach over frequentist regression models as they intended to use the results of the study to inform the priors of the models in a planned follow-up study. 
In total, six models were calculated: one without any predictors to serve as their baseline (which we do not include in this study), one with questionnaire predictors only, two with log data predictors only (one with random intercepts and the other with additional random slopes), and two with all seven predictors (again with random intercepts and additional random slopes respectively). Multilevel models allow the relationship between predictors and outcome variable to vary depending on a grouping factor, which is recommended when there is no valid (e.g., theoretical) reason to assume the relationship remains the same for all values of that factor. For this reason, random slope models were calculated. The grouping factors were the student or the course, depending on the predictors, with the questionnaire scores being associated with the level 2 - grouping factor \emph{student} and the number of clicks on relevant activities being associated with \emph{course}. The remaining predictors were associated with the assignment and thus located on level 1.

\subsubsection{Naive Bayes}
Naive Bayes (NB) utilizes Bayes theorem and assumes that features are independent. In this paper, we use NB for both classification and regression. When applying NB to regression, we first divide the entire range of the outcome variable into $N$ parts and consider each part as a new class. The algorithm then classifies them based on the features and finally reassigns the target values to the classes. The only hyperparameter of NB model is the number of splits, which is determined through cross-validation.  

\subsubsection{K-Nearest Neighbors}
K-Nearest Neighbors (KNN) is a supervised learning algorithm used for both classification and regression. In order to predict the class of a new query point in KNN, we first find its k-nearest neighbors in the training set and then assign the class of the majority of its neighbors to it. The most common metric for measuring the distances and closeness of the points is the Euclidean distance. We can also have nearer neighbors contributing more by assigning a weight proportional to the inverse of distance to the neighbors. When applying KNN to regression problems, the (weighted) average of the neighbors' target values is calculated and assigned to the test point. The hyperparameter $k$ is determined based on cross-validation.

\subsubsection{Radial Basis Function Network}
Conceptually, a neural network is a non-linear function in which the weights must be tuned through training data samples to fit the testing data or unseen observations. RBFN is a simple type of a neural network that has the weights structured in only two layers, a hidden and an output layer. RBFN differ from other neural networks in the type of activation function used in the hidden layer. Each hidden neuron makes use of a radial activation function that calculates the Euclidian distance between each data point and some data centers which are a priori computed through clustering algorithms over the training set. For regression tasks, the activation function of the output neuron is a linear one. When training the RBFN, the weights of hidden and output layers are updated in iterations based on the reinforced error between the predicted and real values. In this sense, the gradient descent algorithm is used to train the weights, where the error rate is an important hyperparameter to be optimized together with the number of hidden nodes and Gaussian parameter.

\subsubsection{Feed-Forward Neural Network}
Feed-forward neural networks (FFNN) differ from RBFN in the sense that the number of hidden layers can be greater than one, the activation function at the level of each hidden node takes simple non-linear representations (e.g., tangent hyperbolic) and at the output layer, the activation function is linear. 
The weights are trained similarly to RBFN in iterations, but the error is back-propagated each time through a greater number of layers, and the weights are updated by using the same gradient descent algorithm. Alongside the error rate, the number of hidden layers and nodes must be decided through cross-validation before training and testing the FFNN structure.

Compared to RBFN, training an FFNN is less complex since a clustering approach is no longer needed. Generally, the most notable disadvantage when training any type of neural network is the lack of criteria when to abort the learning process to prevent over-fitting. In this paper, we propose the following approach to deal with the issue of setting a proper stopping criterion: Out of the training data, a number of validation samples are carefully selected to monitor the performance of the training process. Instead of imposing a fixed stopping criterion, we conducted the training over a large number of iterations and a multi-objective function (to be detailed in Section III.D). The function aimed at balancing the classification performance and minimizing the regression error, and was computed over the validation data. The weights of the neural networks are saved each time a new maximum value is found during the training process. 

\subsubsection{Regression Trees}
The training process of Regression Trees (RT) is known as a binary recursive partitioning, which splits training data into partitions or branches. The algorithms work iteratively and continue splitting each partition into smaller sub-partitions as the training process moves up to each branch. The process of building the tree until each node reaches a specific minimum node size and becomes a terminal node. Once this node is reached, all the responses from all data points are averaged. When testing the RT, each new point follows the split values and variables given by the RT, which was built based on the train data and the predictions are given as a response of the terminal node each new sample ends on. When the minimum node size is 1, the RT can over-fit the training data. In this paper, we determine the minimum node size for each training set based on cross-validation. A very important aspect that concerns the performance of RT is the splitting rule. Standard split criteria (e.g., linear rank statistics, log rank statistics) cannot detect non-linear effects in the outcome variable. To overcome this potential drawback, we use the maximally selected rank statistics for selecting the split point in which splitting variables are compared on the p-value scale \cite{maxstat_wright_2017}. Alongside minimum node size, the lower quantile of the co-variate distribution and the significance level for splitting are all determined through cross-validation.

\subsubsection{Gradient Boosting Machines}
Compared to RT, Gradient Boosting machines (GBM) greedily construct several trees. A new tree is constructed in each round, minimizing the errors given the previously constructed trees. Hence, in each round, the model focuses on the errors made by the previous trees \cite{Friedman:2001:GBM}. The commonly tuned parameters include the number of trees, maximum depth, minimum samples to split the data, learning rate, and the type of loss function. Moreover, by manipulating the sub-sample parameter, a stochastic version of GBM \cite{Friedman:2002:StochasticGBM} uses bootstrapping averaging similar to RF models, when each iteration is trained only on a fraction of the data.

GBM and its variant XGBoost are used in many existing deployed LA systems \cite{Hlosta:2021:ImpactGBM,Ruiperez:2017:earlyMOOC} and were used in winning solutions for predicting student drop-out in the KDD15 competition \cite{KDDcup:2015}. Their high performance is due to their ability to learn from previous mistakes without requiring normalization. As a drawback, the model might suffer from over-fitting, which needs to be overcome by cross-validation and exploring a relatively large number of hyperparameters.

\subsubsection{Random Forest}
In regression tasks, Random Forest (RF) collects the efforts of multiple decision trees and predicts each new data point based on the average responses of all the trees. To decrease the variance of the model without increasing the bias, the trees must be uncorrelated. In this sense, a bootstrapping procedure is applied, and the training subset of each tree is randomly sampled with replacement in the standard proportion of 63.21\% from the overall training set \cite{wright2017}. Moreover, a random subset of features will be considered at each candidate split when training the regression trees. This overcomes a high degree of correlation between trees when some variables are very strong predictors for the delay variable. As a splitting rule, in this paper we employ the principle of ExtraTrees in which random cut-points are selected in the top-down splitting process \cite{wright2017}. For each possible splitting variable, a number of random cut-points are generated and the one with the highest decrease of impurity is selected to split the node. Therefore, the RF algorithm adjusts four hyperparameters based on cross-validation: the number of regression trees, the minimum node size for all trees, the number of possible variable splits, and the number of cut values that are randomly generated in the range of min/max values for each possible split variable.

RF is less prone to over-fitting compared to simple RT or GBM. However, the RF algorithm may change considerably by a small change in the training data. Moreover, the RF training time increases considerably in larger data sets, especially when computing the optimal cut-point locally for each feature (i.e., based on log-rank statistics, maximal rank statistics, etc.).

\subsubsection{Support Vector Regression}
Being originally proposed for binary classification problems, support vector machines aim to find the best hyperplane that separates a given set of data points. Training points are mapped in space so that the width of the gap or the soft margin between the two categories would be maximized. Test points are mapped in this space and predicted as being part of one of such categories depending on which side of the gap they belong. In practice, the soft margin that separates the two categories of data points can be controlled based on parameter $C$. When $C$ is relatively small, a larger separation margin between the classes will be used at the price of higher rate of misclassifications. When $C$ is higher, the separation gap gets smoother and the rate of misclassifications is lower. A lower misclassification rate increases the risk of over-fitting the training data that can lead to a much higher rate of misclassification for new data points. An optimal value of soft-margin hyperparameters must be found on training data to fit the testing data.

The accuracy of classification depends on the shape of the separation hyperplane that can be set through the kernel function. In this paper, we test different types of kernel functions (linear, polynomial, radial basis function, radial basis function with vector subtraction, hyperbolic tangent), that can be used to best fit the collected data. Moreover, some of these kernel functions must be parameterized before properly training and testing the machine. In regression tasks, the goal of Support Vector Regression (SVR) is comparable to classification tasks with the amendment that a certain error is tolerated. In this paper, we optimize the soft margin parameter, the kernel function and error tolerance through cross-validation.

\vspace{-0.1in}
\subsection{Proposed Cross-Validation Measure}
In order to optimize the hyperparameters of the implemented ML algorithms (NB, KNN, RBFN, FFNN, RT, RF, GBM, SVR), we employed a cross-validation procedure based on the training sets. Each ML model defines its own grid of hyperparameter configuration that falls in predefined ranges. Each possible configuration of hyperparameters from the grid is then evaluated based on the training data. Finally, the trained ML model with the highest evaluation outcome is employed to predict new data examples from the testing set.

The algorithm splits the training set in $K$ number of sub-partitions. Then, the $K$-fold cross-validation trains the ML model iteratively for each hyperparameter configuration band on $K-1$ sub-partitions and the results are validated based on the remaining sub-partition every time. For each configuration, the validation results are averaged over $K$ number of partitions. Once all hyperparameter configurations are evaluated, the $K$-fold cross-validation algorithm selects the scheme with the highest validation outcome. The same procedure is repeated for each  ML algorithm with its predefined grid search of hyperparameter configurations. The validation outcome can be measured by monitoring the regression error (e.g., mean absolute error). However, in our prediction task, the selected hyperparameters should also improve the classification performance.

Moreover, the ratio of early and late submissions in the training data is often unbalanced. Therefore, we propose a novel multi-objective function that measures the validation performance of each hyperparameter configuration on each split of data set. Based on this proposed function, the cross-validation selects the hyperparameter configuration that minimizes the regression error and maximizes the classification performance between the two classes (delay and timely submission).

Given the function $f$ that must be trained by each ML model to best predict delay, we denote by $\hat{y} = f(\mathbf{x})$ the current estimate of prediction function $f$, where $\mathbf{x}$ is a test data point. In order to evaluate the prediction performance, the ground truth values of delay $y$ should be known in advance. In this study, delay $y$ is measured as the difference between the submission of an assignment and its deadline in days. The deadline is met when $y \leq 0$ (timely submission), and the tasks are submitted after the deadline when $y > 0$ (delay). When a new data sample from the test set is predicted, the decision can take one of the following forms: \textit{a)} true positive ($TP$): $\hat{y} > 0$ and $y > 0$; \textit{b)} false positive ($FP$): $\hat{y} > 0$ and $y \leq 0$; \textit{c)} true negative ($TN$): $\hat{y} \leq 0$ and $y \leq 0$; \textit{d)} false negative ($FN$): $\hat{y} \leq 0$ and $y > 0$. Once all test samples are predicted, the classification performance is evaluated based on $TP$, $FP$, $TN$ and $FN$ indicators which are summed over the number of test data examples. To measure the ML model performance when dilatory behavior is correctly detected, the F1-score function can be employed as follows:
\begin{equation}
\label{eq:1}
F_{tp} = \frac{2TP}{2TP + FP + FN}
\vspace{-0.05in}
\end{equation}
where a higher $F_{tp}$ value implies a better performance when predicting delay. In unbalanced data sets, $F_{tp}$ does not give any measure of how well the prediction performance is for the timely submission class. In this case, a similar version of F1-score measure can be computed to monitor the ratio of $TN$ over the number of false predictions:
\begin{equation}
\label{eq:2}
F_{tn} = \frac{2TN}{2TN + FP + FN}
\vspace{-0.05in}
\end{equation}
By maximizing both $F_{tp}$ and $F_{tn}$ in cross-validation, the hyperparameters will be selected according to the best prediction balance between the two classes. To measure how close the predicted value $\hat{y}$ is from its delay pattern $\hat{y}$, we measure the Mean Absolute Error (MAE) $E$ given by:
\begin{equation}
\label{eq:3}
E = \frac{1}{T}\sum_{i=1}^T \frac{|\hat{y}_i - y_i|}{y_{max}}
\vspace{-0.05in}
\end{equation}
where $T$ is the number of samples in each data set split and $y_{max}$ is the maximum absolute value of delay. Considering (1), (2) and (3), the multi-objective function to be maximized during the cross-validation process becomes:
\begin{equation}
\label{eq:4}
G = \frac{(1-E) + F_{tp} + F_{tn}}{3}
\vspace{-0.05in}
\end{equation}
The configuration of hyperparameters able to maximize (4) will be selected to train the ML model for the entire training set and the prediction performance is analyzed in Section IV. Each ML algorithm follows the same cross-validation principle when tuning the corresponding hyperparameters.

\vspace{-0.1in}
\subsection{Analyses}
The baseline models (i.e., the Bayesian multilevel models) were calculated with the R package \textit{brms} \cite{Buerkner2017brms} and the NB, KNN and GBM algorithms were employed using the Python \textit{scikit-learn} package \cite{pedregosa2011scikit}. The remaining ML approaches (FFNN, RBFN, RT, RF, SVR) were deployed in C/C++ using dedicated functions for cross-validation, training and testing. For the \texttt{obj} and \texttt{comb} predictors, the Bayesian models included random intercepts (BA-RI) and random slopes (BA-RS) due to the nested structure of the data, which was not the case for the \texttt{subj} predictors. The implementation of RT and RF was based on RANGER C++ packages which are publicly available \cite{wright2017}. The proposed SVR ML algorithm followed the regression model from \cite{parella2007} with five different types of kernel functions: linear (SVR-LIN), polynomial (SVR-POL), tangent hyperbolic (SVR-TAH), radial-basis function (SVR-RBF) and with the vector subtraction function (SVR-VS). All these variants of SVR model are cross-validated, trained, tested and compared to the other ML algorithms. The FFNN model we used was previously employed for the optimization of video quality in remote education \cite{comsa_education_2021}, while the RBFN model was also used to classify high dimensional vectors in radio communications systems \cite{comsa_thesis_2014}. 

The complexity of cross-validation processes depends on the grid size, which differs from one ML algorithm to another. A special case in cross-validation is represented by RBFN, where the number of hidden nodes is equivalent to the number of data centers computed for each data set separately based on the clustering analysis. The clustering process is conducted before cross-validation by employing a heuristic algorithm that iteratively combines a classical k-means algorithm for a more precise calculation of centers with an algorithm that uses the random swapping of data centers from the available data set to enhance the searching time of globally optimal solutions \cite{comsa_thesis_2014}. Based on the clustering algorithm, we determined the number of optimal clusters that characterizes each training set by employing an additional algorithm to calculate the Silhouettes Index (SI) \cite{si_index}. The SI index interprets and validates the consistency of the clusters for each training set, and provides a measure of how well each data point is matched to its own cluster compared to the neighboring clusters. Higher SI values (max value of 1) denote that the data points are very well suited to their clusters, while lower values (min value of -1) indicate that the clustering configuration may have too many or too few clusters. 
In general, when a higher number of clusters is obtained through the computation of SI index, then the data set is much better represented when training the RBFN model and a higher prediction performance is expected.

\begin{table*}
\centering
\small
\caption{Comparison of $G$-scores between different predictor types and ML approaches obtained in cross-validation and prediction}
\begin{tabular}{cccc|ccc}\hline
\label{table04}
 & \multicolumn{3}{c}{\textbf{Cross-validation}} & \multicolumn{3}{c}{\textbf{Test}}\\
  & \multicolumn{3}{c}{\textbf{Mean $G$-Score (SD)}} & \multicolumn{3}{c}{\textbf{Mean $G$-Score (SD)}}\\
\textbf{ML Alg.} & \texttt{subj} & \texttt{obj} & \texttt{comb} & \texttt{subj} & \texttt{obj} & \texttt{comb} \\
\hline
\textbf{NB} & 0.6457 (0.0084) & 0.6845 (0.013) & 0.6961 (0.0048) & 0.6375 (0.021) & 0.6628 (0.0267) & 0.681 (0.019)  \\
\textbf{KNN} & 0.7071 (0.0068) & 0.7482 (0.0039) & 0.7525 (0.0069)  & 0.6758 (0.0222) & 0.7279 (0.0132)  & 0.7319 (0.019) \\
\textbf{RBFN} & \cellcolor{inchworm} 0.7401 (0.0051) & \cellcolor{inchworm} 0.7707 (0.0072)  & 0.7688 (0.0054) & 0.6778 (0.0179) & 0.701 (0.0218) & 0.7214 (0.0204) \\
\textbf{FFNN} & 0.7143 (0.012) & 0.7631 (0.0102) & 0.7673 (0.008)  & 0.6695 (0.0368)  & 0.6957 (0.0316) & 0.71884 (0.0223)  \\
\textbf{RT} & 0.6772 (0.0106) & 0.7491 (0.0046) & 0.7478 (0.0047) & 0.6811 (0.0251) & 0.7126 (0.025) & 0.7181 (0.035) \\
\textbf{RF} & 0.7081 (0.008) & 0.755 (0.0057) & 0.7685 (0.0054) & 0.7064 (0.0153) & \cellcolor{inchworm} 0.7457 (0.018) & \cellcolor{inchworm} 0.7626 (0.015)  \\
\textbf{GBM} & 0.70686 (0.01143) & 0.76012 (0.00919) & \cellcolor{inchworm} 0.7713 (0.0071) & \cellcolor{inchworm} 0.71624 (0.025) & 0.73166 (0.021) & 0.7442 (0.0271) \\
\textbf{SVR-LIN} & 0.6785 (0.0071) & 0.7253 (0.0083) & 0.7275 (0.0069) & 0.6591 (0.0422) & 0.6888 (0.0264) & 0.7134 (0.019) \\
\textbf{SVR-POL} & 0.6975 (0.0056) & 0.7348 (0.0041) & 0.7413 (0.0055) & 0.6835 (0.0312) & 0.6934 (0.0273) & 0.715 (0.023) \\
\textbf{SVR-TAH} & 0.6833 (0.01) & 0.7305 (0.0072) & 0.7331 (0.0062) & 0.6544 (0.0318) & 0.6916 (0.0334) & 0.708 (0.024) \\
\textbf{SVR-RBF} & 0.7208 (0.0094) & 0.7406 (0.0059) & 0.7536 (0.0077) & 0.7129 (0.0189) & 0.7288 (0.026) & 0.7538 (0.018) \\
\textbf{SVR-VS} & 0.7208 (0.0084) & 0.7608 (0.0045) & 0.7631 (0.0048) & 0.7145 (0.0175) & 0.7364 (0.0257) & 0.7589 (0.018) \\
 \hline
\end{tabular}
\end{table*} 

We evaluated the performance of the ML algorithms for each type of predictor based on classification metrics and regression errors ($G$-Scores for the former and MAE for the latter). 
As classification measures, we considered the $F_{tp}$ and $F_{tn}$ scores in the computation of the multi-objective $G$ function. 
When training data is unbalanced, we recommend the use of both scores to find the best configuration of hyperparameters that can balance precision and robustness in both directions. This is the case for our original data set, which is unbalanced in favor of the timely-submission class, meaning that 67\% of assignment data points were classified as timely submissions ($y<0$), and the remaining 33\% as delay ($y>0$). Therefore, we included the following metrics: Positive Predicted Value $(PPV = TP/(TP+FP))$, which is the precision of detecting delay; and the True Positive Rate $(TPR = TP/(TP+FN))$ that indicates how well delay can be predicted without negatively affecting the prediction of timely submissions (sensitivity or recall). However, when data is perfectly balanced, the use of $F_{tp}$ and $F_{tn}$ is not needed. In this case, other classification metrics could be used instead, including accuracy (ACC) as the ratio of the correct and total predictions, and the Matthews Correlation Coefficient (MCC) as a metric of difference between correct and wrong predictions. Generally, the MCC is reportedly more informative compared to other measures \cite{chicco2020advantages, chicco2017ten}. We included ACC and the MCC for the sake of completion. 

\section{Results}
The aim of this section is to evaluate the performance of ML algorithms in regard to cross-validation and prediction. First, we measured and evaluated the multi-objective function $G$-Score for each algorithm and type of predictor in the cross-validation process. This was followed by a calculation of the same $G$-Scores for the test data sets. These results were then compared to those of the cross-validation (RQ1). RQ2 is then addressed by comparing the ML approaches for each type of predictor in more detail while considering the classification performance indicators and the mean absolute error. Finally, we address RQ3 by evaluating how high the predictive performance is when considering the \texttt{obj} and \texttt{comb} predictors compared to predictions based solely on \texttt{subj} variables.

As explained above, the ML algorithms were cross-validated, trained and tested separately for ten randomized subsets to account for the discovered unbalance. The reported results include the mean and SD values over the ten subsets for each of the three types of predictors. 

\subsection{Evaluation in Cross-Validation}
In this paper, we employed a 4-fold cross-validation scheme for all of the ML algorithms, in which all training sets were divided in four subsets with 222 elements each. 
The same cross-validation procedure was conducted for the three types of 
predictors and each ML algorithm, resulting in a total of 360 different cross-validation processes.

\subsubsection{Mean G-Score in Cross-Validation}
To evaluate the performance of the ML algorithms during cross-validation, we computed the mean $G$-score and SD over ten data sets for each of the three predictor types. The values exposed in Table II for each ML algorithm are the maximum $G$-scores of the best configuration of hyperparameters obtained through the cross-validation averaged over ten training data sets.

We do not report hyperparameters for the Bayesian multilevel models since we selected non-informative priors, as was the case in the previous study \cite{Imhof2021}. Looking at the \texttt{subj} and \texttt{obj} predictors, RBFN obtained the highest scores of $0.74$ and $0.771$, respectively. The performance regarding the validation sets was higher compared to the other ML algorithms since the data centers obtained through the proposed clustering approach cover the entire training set. For the \texttt{comb} predictors, RBFN, RF and GBM obtained comparable $G$-scores. When using SVR, the LIN and TAH kernel functions are not suitable options to predict delay when employing cross-validation over the training sets. The non-linear kernels (POL, RBF, VS) can fit the validation sets better in the training $K-1$ sub-partitions. 
In the next step, we verified if the same prediction trend in cross-validation is followed when predicting new examples from the test sets.

\begin{table*}
\centering
\small
\caption{Intra-model comparison between ML approaches for \texttt{subj} predictors}
\begin{tabular}{cccccccccc}\hline
\label{table04}
\textbf{ML Alg.} & PPV (SD) [\%] & TPR (SD) [\%] & $F_{tp}$ (SD) [\%] & $F_{tn}$ (SD) [\%] & MCC (SD) [\%] & ACC (SD) [\%] & MAE (SD) \\
\hline
\textbf{BA} & 36.19 (3.61) & 32.59 (8.16) & 33.66 (5.59) & 71.33 (2.69) & 5.91 (4.22) & 60.14 (2.39) & 9.19 (0.99) \\
\textbf{NB}	& 41.1 (4.4) &	41.6 (8.5) & 40.9 (5.1) &	73.3 (2.9) & 14.8 (5.4) & 63.4 (2.6) & 30.38 (2.47) \\
\textbf{KNN} &	39.44 (5.32) & 44.84 (8.62)	& 41.68 (5.84) & 70.09 (2.78)	& 12.4 (7.81)	& 60.55 (3.27) & 11.94 (2.57) \\
\textbf{RBFN} & 40.18 (3.28) & 42.66 (7.65) & 40.9 (3.37) & 71.06  (3.039) & 12.73 (3.02) & 61.3 (2.4) & 11.39 (1.35) \\
\textbf{FFNN} &	40.46 (5.45) & 37.1 (14.06) & 37.04 (5.33) & 71.69 (6.49) &	11.21 (5.88) & 61.62 (5.09) & 10.44 (1.03) \\
\textbf{RT}	& 39.35	(4.03)	& \cellcolor{inchworm} 46.01 (8.97)	& 42.19	(5.61) & 69.76 (2.81) & 12.68 (6.57) & 60.45 (6.48) & 10.09	(1.15) \\
\textbf{RF}	& 45.82	(4.15) & 45.86	(5.83)	&  45.43 (2.68) & 74.28 (2.74)	& 20.22	(3.09) &  65.18	(2.42) & 10.31	(1.1) \\
\textbf{GBM} & 47.69 (4.75) & 43.94 (7.18) & \cellcolor{inchworm} 45.56	(5.4) & 76.2 (1.77)	& 22.07 (7.04) & 66.89 (2.67) & 9.11 (0.95) \\
\textbf{SVR-LIN} &  42.46 (7.88) & 25.96 (12.69) & 29.50 (9.81) &  75.42 (3.52) & 9.66 (4.35) &  64.05 (2.92) & 9.54 (0.89) \\
\textbf{SVR-POL} &  42.02 (2.79) &  36.47 (10.14) & 38.31 (5.59) & 73.80 (4.29)	& 13.30	(3.72)	& 63.65 (3.19) & 9.34 (0.96) \\
\textbf{SVR-TAH} & 39.84 (7.95) & 22.39 (8.85) & 27.52 (7.66) &  75.89 (2.62) & 7.28 (6.33) & 64.05 (2.78) & 9.37 (0.95) \\
\textbf{SVR-RBF} & 50.76 (4.45) & 36.56 (5.99) & 42.24 (4.88) & 78.30 (1.62) & 22.14 (5.11) & 68.51 (2.02) & \cellcolor{inchworm} 8.82 (1.09) \\
\textbf{SVR-VS}	& \cellcolor{inchworm} 51.22 (4.29) & 36.57 (5.46) & 42.51 (4.7) & \cellcolor{inchworm} 78.54 (1.31) & \cellcolor{inchworm} 22.63 (5.05) & \cellcolor{inchworm} 68.78 (1.83) & 8.84 (1.03) \\
 \hline
\end{tabular}
\end{table*} 

\begin{table*}
\centering
\small
\caption{Intra-model comparison between ML approaches for \texttt{obj} predictors}
\begin{tabular}{cccccccccc}\hline
\label{table04}
\textbf{ML Alg.} & PPV (SD) [\%] & TPR (SD) [\%] & $F_{tp}$ (SD) [\%] & $F_{tn}$ (SD) [\%] & MCC (SD) [\%] & ACC (SD) [\%] & MAE (SD) \\
\hline
\textbf{BA-RI} & 47.97 (5.11) & 57.53 (4.78) & 52.23 (4.51) & 74.17 (3.02) & 27.13 (6.51) & 66.53 (3.32) & 8.57 (0.68) \\
\textbf{BA-RS} & \cellcolor{inchworm} 51.02 (3.52) & \cellcolor{inchworm} 60.88 (6.75) & \cellcolor{inchworm} 55.41 (4.37) & 76.14 (2.32) & \cellcolor{inchworm} 32.29	(5.4) &	\cellcolor{inchworm} 69.01 (2.38) & 7.57 (0.63) \\
\textbf{NB} & 40.6 (6.1) & 29.2 (7.8) & 33.7 (7.2) & 76.3 (1.7) & 11.5 (6.5) & 65.2 (2.0) & 14.75 (1.19) \\
\textbf{KNN} & 49.1 (1.77) & 48.6 (5.24) & 48.69 (2.94) & 76.26 (1.68) & 25.15 (3.27) & 67.62 (1.69) & 8.72 (0.88) \\
\textbf{RBFN} & 43.49 (4.34) & 52.59 (9.99) & 46.79 (3.16) & 70.72 (4.36) & 19.45 (3.08) & 62.52 (3.25) & 9.64 (1.26) \\
\textbf{FFNN} & 43.55 (3.8) & 45.29 (11.01) & 43.43 (5.74) & 72.54 (4.38) & 17.22 (4.44) & 63.47 (3.46) & 9.57 (0.87) \\
\textbf{RT} & 48.39 (4.25) & 40.76 (8.06) &	43.8 (5.73) & 76.71	(2.62) & 21.39 (5.29) & 67.25 (2.62) & 8.92 (1.09) \\
\textbf{RF} & 50.36 (4.31) & 57.12 (5.09) & 53.48 (4.4) & 76.16 (1.68) & 29.98 (5.7) & 68.51 (2.25) & 7.88 (0.79) \\
\textbf{GBM} & 50.7 (5.04) & 45.13 (7.04) & 47.51 (5.46) & \cellcolor{inchworm} 77.53 (1.74) & 25.52 (6.12) & 68.61 (2.24) & \cellcolor{inchworm} 7.34 (0.86) \\
\textbf{SVR-LIN} & 43.78 (5.21) & 35.72 (9.61) & 38.48 (6.38) & 75.08 (2.27) & 14.99 (5.99) & 64.69 (2.47) & 9.13 (0.93) \\
\textbf{SVR-POL} & 43.13 (6.54) & 38.39	(8.43) & 40.23 (6.24) & 74.32 (2.62) & 15.21 (7.79) & 64.19 (3.15) & 8.66 (0.89) \\
\textbf{SVR-TAH} & 47.18 (5.35) & 35.44 (13.82) & 38.63 (8.16) & 76.26 (3.55) & 17.77 (4.93) & 66.26 (2.78) & 9.81 (2.22) \\
\textbf{SVR-RBF} & 46.47 (5.54) & 41.09 (8.14) & 43.23 (5.96) & 75.76 (2.46) & 19.64 (6.31) & 66.17 (2.58) & 8.39	(0.85)  \\
\textbf{SVR-VS}	 & 49.73 (4.57) & 51.81 (9.62) & 50.52 (6.82) & 76.50 (1.59) & 27.35 (7.06)	& 68.29 (2.07) & 8.09	(0.91) \\
 \hline
\end{tabular}
\vspace{-0.05in}
\end{table*} 

\subsubsection{Mean G-Score in Test Data Sets}
The hyperparameters obtained through cross-validation were used to train each ML algorithm by using the entire training sets (K folds) for 
each type of predictor. 
Their performance is then evaluated 
in the test sets for all ML algorithms based on $G$-Scores. 
In Table II (right side), we present the mean $G$-scores and SD values calculated over the ten test data sets for each type of predictor. 
In case of the \texttt{subj} predictors, the GBM and SVR-VS models outperformed the others with mean $G$-scores higher than $0.71$. 
When analyzing the \texttt{obj} predictors, the best performance of $G=0.75$ was achieved by the baseline Bayesian models with random slopes. When combining both types of predictors, RF and SVR-VS remained the best options with a mean $G$-score higher than $0.76$. 

\subsubsection{Mean G-Score Comparison} 
In order to validate the configuration of hyperparameters selected for each ML algorithm, the $G$-scores were compared between cross-validation and testing. The most successful ML approach would have the highest $G$-score among all candidates and all types of predictors with the smallest performance deprecation between cross-validation and testing. NB and KNN both had a rather small difference in $G$-scores in the validation and testing sets, but the performance level is much lower compared to the other ML approaches. RT and GBM are well known for over-fitting the training data, which can also be observed when looking at their $G$-scores. SVR is sensitive to the training data, and the performance of delay prediction depends very much on the selected configuration of hyperparameters. This explains the 3\% deprecation between validation and test sets for all types of predictors, especially for the case of kernels that use the linear and tangent hyperbolic functions. When analyzing the performance of RBFN and FFNN in the validation and test data sets, we observed the highest degradation in performance. This aspect is explained by the fact that both RBFN and FFNN were trained on 60\% of the data, while the remaining 20\% were used for the stopping criteria. Also, the $G$-scores reported in cross-validation are the best values that could be found while training in each split. RF was the most stable ML algorithm for all three types of predictors, since the levels of test $G$-scores were very close to those scores obtained in cross-validation. 


\subsection{Intra-Model Comparisons}
In this section, we provide a comprehensive report to compare the employed ML algorithms for each type of predictor (intra-model comparison). Thus, we analyse the mean and SD of the PPV, TPR, $F_{tp}$, $F_{tn}$, MCC, ACC, and MAE performance indicators, where the best values of these indicators are highlighted in green. To answer RQ2, the goal of this section is to determine the most successful ML algorithm for each type of predictor that would maximize the $F_{tp}$ measure and minimize the loss in $F_{tn}$ and regression error (MAE).

\subsubsection{Subjective Predictors}
In Table III we present the classification and regression performance metrics (mean and SD) for each ML algorithm and calculated over the data sets with \texttt{subj} predictors. When looking at the PPV metric, we obtained the highest amount of correct predictions for the delay class when using SVR with RBF and VS kernels. However, the robustness of these models in predicting the timely submission class is deprecated more than 15\% when compared to PPV. The best trade-off between PPV and TPR was observed with the RF and GBM models. When computing the $F_{tp}$ score, GBM performed slightly better than RF with 45.5\%. However, the SD of the RF model is reduced by half when compared to GBM, meaning the RF classifier is less sensitive to the type of data set which is used to train the model. The $F_{tn}$ score was also calculated in Table III to measure the trade-off between the precision for timely submission and the robustness to delay. Since the false predictions of delay have a greater impact than the false predictions of timely submissions due to the unbalance of the data sets, the SVR model with the RBF and VS kernels achieved the highest $F_{tn}$ scores. By monitoring the MCC and ACC classification metrics, it can be concluded that the absolute difference between the false predictions of delay and timely submission was higher for the SVR model (with RBF and VS kernel functions) than for the RF and GBM models. Based on the results collected in Table III, a precision to detect delay of nearly 48\% and a classification accuracy of about 67\% can be obtained with the GBM model when exclusively considering the \texttt{subj} predictors. When computing the regression error, SVR model with RBF and VS kernels performed slightly better than GBM. However, we recommend the use of the GBM model when predicting the \texttt{subj} predictors due to the best trade-off between regression error, $F_{tp}$ and $F_{tn}$.

\begin{table*}
\centering
\small
\caption{Intra-model comparison between ML approaches for \texttt{comb} predictors}
\begin{tabular}{cccccccccc}\hline
\label{table04}
\textbf{ML Alg.} & PPV (SD) [\%] & TPR (SD) [\%] & $F_{tp}$ (SD) [\%] & $F_{tn}$ (SD) [\%] & MCC (SD) [\%] & ACC (SD) [\%] & MAE (SD) \\
\hline
\textbf{BA-RI} & 49.72 (5.17) & 59.94 (3.82) & 54.25 (4.14)	& 75.14	(2.82) & 30.20 (5.85) & 67.84 (3.14) & 8.56	(0.68)  \\
\textbf{BA-RS} & 51.04 (2.65) & 61.08 (5.43) & 55.55 (3.49) & 76.19 (1.6) & 32.43 (4.09) & 69.05 (1.62) & 7.61 (0.66)  \\
\textbf{NB} & 42 (5.3) & 40.9 (5.5) & 41.2 (4.1) & 74.4	(3.) & 15.9 (6.1) & 64.5 (3.3) & 15.12 (1.74) \\
\textbf{KNN} & 49.34 (5.22) & 51.8 (4.72) & 50.41 (4.19) & 67.67 (2.19) & 26.6 (5.92) & 67.67 (2.64) & 9.01 (0.89)  \\
\textbf{RBFN} & 46.28 (5.02) & 59.69 (6.79) & 51.69 (3.37) & 71.95 (3.69) & 25.58 (4.68) & 64.64 (3.35) & 9.56 (1.22)  \\
\textbf{FFNN} & 47.72 (4.45) & 47.28 (7.54) & 47.21 (5.34) & 75.59 (2.16) & 23.15 (5.36) & 66.76 (2.17) & 9.46 (1.07)  \\
\textbf{RT} & 48.87 (7.61) & 49.42 (10.77) & 47.92 (5.09) & 74.41 (6.28) & 24.02 (6.63)	& 66.26 (5.31) & 9.13	 (1.24) \\
\textbf{RF} & 52.75 (3.99) & \cellcolor{inchworm} 62.22 (3.8) & \cellcolor{inchworm} 57.05 (3.58) & 77.23 (1.61) & \cellcolor{inchworm} 34.86 (4.6) & 70.27 (1.98) & 7.27 (0.82)  \\
\textbf{GBM} & 51.60 (5.47) & 51.35 (8.96) & 51.15 (6.58) & 77.45 (2.19) & 28.98 (7.71) & 69.23 (2.94) & \cellcolor{inchworm} 7.05 (0.78)  \\
\textbf{SVR-LIN} & 47.28 (5.28)	& 43.79	(4.78) & 45.24 (3.85) & 75.66 (2.43) & 21.26 (5.71) & 66.35 (2.84) & 9.1	(0.73)  \\
\textbf{SVR-POL} & 47.03 (3.59) & 44.86 (8.40) & 45.55 (5.39) & 75.59 (2.22) & 21.63 (5.94) & 66.39 (2.45) & 8.82	(0.91)  \\
\textbf{SVR-TAH} & 47.84 (3.28)	& 38.94 (10.18) & 42.27 (6.43) & 76.87 (1.48) & 20.46 (5.28) & 67.12 (1.51) & 8.93 (0.87)  \\
\textbf{SVR-RBF} & \cellcolor{inchworm} 53.83 (5.27) & 54.67 (7.06) & 53.92 (4.31) & \cellcolor{inchworm} 78.23 (1.88) & 32.57 (5.44) & \cellcolor{inchworm} 70.5 (2.2) & 7.97 (0.89)  \\
\textbf{SVR-VS} & 53.12 (4.73) & 58.82 (4.54) & 55.72 (3.99) & 77.66 (2.12) & 33.69 (5.86) & 70.32 (2.66) & 7.57 (0.99) \\
\hline
\end{tabular}
\end{table*} 

\subsubsection{Objective Predictors}
In Table IV, we present the classification and regression performance metrics (mean and SD) obtained when predicting delay with \texttt{obj} predictors. When evaluating the predictive performance of the delay class, the Bayesian model with random slopes (BA-RS) provided the best results among all models with the following metrics: a) the precision (PPV = 51.02\%) when predicting delay; b) robustness (TPR = 60.88\%) to predict timely submission; c) trade-off between PPV and TPR ($F_{tp}=55.41\%$). Also, BA-RS was the best option when measuring the MCC (32.29\%) and accuracy (ACC = 69.01\%). On average, the GBM model achieved the highest $F_{tn}$ score of 77.53\% due to higher robustness to predict delay. Also, its error was lower compared to the other models. When looking at the trade-off between $F_{tp}$, $F_{tn}$ and mean absolute error, BA-RS was the best option to predict delay with \texttt{obj} predictors.

\subsubsection{Subjective and Objective Predictors combined}
When looking at the performance of the ML algorithms for the \texttt{comb} predictors in Table V, we observed that the SVR-RBF performed better in precision (with about 2\%) but with lower robustness value (with more than 7\%) when compared to the RF model. This explains the larger gain obtained by the RF algorithm when measuring the $F_{tp}$ score of about 57\%. The same trend can be observed when comparing the SVR-VS and RF models. When measuring the $F_{tn}$ scores, SVR with RBF kernel provided the best results due to better robustness of the model to predict delay. By comparing the RF, SVR-VS, and SVR-RBF models, we noted that higher $F_{tp}$ scores involve higher MCC levels, while higher $F_{tn}$ scores align with higher accuracy values. When calculating the average regression error, GBM and RF models outperformed the other ML models. Although SVR-RBF and SVR-VS provided better precision (PPV), the RF model remained the best option to be employed for the \texttt{comb} predictors due to the best trade-off between precision and robustness to delay and timely submission classes, and a low regression error.

\begin{table*}
\centering
\small
\caption{Inter-model comparison between the highest-performing ML approach for each type of predictor: GBM (\texttt{subj}), BA-RS (\texttt{obj}) and RF (\texttt{comb})}
\begin{tabular}{ccccccccc}\hline
\textbf{Pred type} 
& PPV (SD) [\%] & TPR (SD) [\%] & $F_{tp}$ (SD) [\%] & $F_{tn}$ (SD) [\%] & MCC (SD) [\%] & ACC (SD) [\%] & MAE (SD) \\
\hline
\texttt{subj} 
& 47.69 (4.75) & 43.94 (7.18) & 45.56	(5.4) & 76.2 (1.77)	& 22.07 (7.04) & 66.89 (2.67) & 9.11 (0.95) \\
\texttt{obj} 
& 51.02 (3.52) & 60.88 (6.75) & 55.41 (4.37) & 76.14 (2.32) & 32.29	(5.4) & 69.01 (2.38) & 7.57 (0.63) \\
\texttt{comb} 
& 52.75 (3.99) & 62.22 (3.8) & 57.05 (3.58) & 77.23 (1.61) & 34.86 (4.6) & 70.27 (1.98) & 7.27 (0.82)  \\
\hline
\end{tabular}
\end{table*} 


\subsection{Inter-Model Comparison}
In this section, we compare the predictive performance between each type of predictor to answer RQ3 by taking into account the results displayed in Tables II-V.

When analyzing the performance of the mean $G$-scores in both the cross-validation and testing stages (Table II), we concluded that the \texttt{obj} predictors are more informative than the \texttt{subj} predictors when predicting delay in assignments. However, a performance gain can be achieved by combining both types of predictors. When analyzing the best $G$-scores in Table II (on the right side), we observe a prediction gain of 4\% when comparing the prediction with the \texttt{obj} and \texttt{subj} predictors, and a gain of 1\% when predicting with the \texttt{comb} predictors compared to the \texttt{obj} predictors alone. By combining both \texttt{subj} and \texttt{obj} predictors, almost all ML candidates benefit from the perspective of both classification and regression metrics as shown in Table V. For example, the SVR model with all types of kernel functions obtained higher precision (PPV) and robustness (TPR) when involving the \texttt{comb} predictors compared to the \texttt{subj} or \texttt{obj} predictors alone. The same performance gain can be observed when measuring MCC, ACC, or MAE. Another concluding example in this sense is the RF approach that enhanced its $F_{tp}$ score from 45.43\% for \texttt{subj} predictors to 53.48\% for \texttt{obj} data, and when combining both, the performance was higher than 57\%. An explanation of this gain is given by the importance of the \texttt{subj} predictors that changes when the \texttt{obj} predictors are added. In this sense, we computed the Gini importance index or the mean decrease in impurity that gives the feature importance when regression trees are employed \cite{gini_index_2021}. When training the RF model, we make the following observations based on the importance index: \textit{a)} in case of \texttt{obj} predictors, the interval between the start of a block and the first click on an assignment is the most important predictor, followed by the number of clicks on an assignment and the number of clicks on relevant activities; \textit{b)} when training with \texttt{subj} predictors, APS is the most informative, followed by APSS, SDLS, and GASE; \textit{c)} by combining both types of predictors, the most important variables are the \texttt{obj} predictors in the same order as above in \textit{a)}, followed by the \texttt{subj} predictors in the order as before (\textit{b}), with the only difference that the GASE variable gained a higher importance than APS. While APS and APSS are more predictive in general, when combined with \texttt{obj} predictors, they do not add any new information. Therefore, the GASE variable could enhance the precision and accuracy above 52\% and 70\% respectively when combining both sets of predictors.

As observed in Tables III-IV, the highest performing ML approach differs from one type of predictor to another when balancing the performance between $F_{tp}$ and $F_{tn}$ scores and the regression error. GBM is recommended to be used for \texttt{subj} data due to a better trade-off between the indicators above. 
When using the \texttt{obj} predictors only in the form exposed in Table I, the BA-RS approach was the best option to be employed. However, for combined predictors, the RF method outperformed other approaches due to a higher importance of GASE feature among other \texttt{subj} ones. In Table VI, we summarize the results from Tables III-IV and present the inter-model comparison between the highest performing ML approach for each type of predictor. The combined predictors brought a performance gain of more than 1\% when compared to \texttt{obj} data sets and more than 11\% when compared to \texttt{subj} predictors. The regression error was the lowest when predicting with the \texttt{comb} data sets while the deprecation in $F_{tn}$ was negligible between different types of predictors. The highest accuracy value of 70.23\% was achieved when combining the predictors, with a slight depreciation of about 1\% for \texttt{obj} and 3\% for \texttt{subj} predictors. The same trend was observed when monitoring the MCC indicator with a much larger performance gap of 10\%. 

\section{Discussion}
The aim of this study was to expand the findings in \cite{Imhof2021} by employing multiple ML algorithms to determine which one delivered the best predictions of delay based on objective and subjective variables. First, due to the unbalanced nature of the data (meaning there were twice as many timely submissions as there was delay), we needed to account for it in our approach to optimize each of the ML algorithms. We achieved this by a cross-validation procedure we conducted based on a novel multi-objective function, the $G$-score, which involves a trade-off between positive and negative $F$1-scores and regression error. We measured the performance degradation of $G$-score between test and cross-validation to verify the authenticity of the selected hyper-parameters for each ML algorithm. Except for RBFN and FFNN, all other approaches provided a low degradation in $G$-score. We concluded that RBFN and FFNN need more data to cope with such a performance gap. Among the ML algorithms in our pool, RF obtained the lowest degradation in performance, meaning that cross-validation was highly efficient (RQ1).

We then identified the best prediction performances for each of the types of predictors (\texttt{subj}, \texttt{obj}, \texttt{comb}) by measuring the trade-off between $F_{tp}$, $F_{tn}$, and MAE (RQ2). GBM turned out to be the best approach for subjective predictors, BA-RS for the objective predictors, and RF for the combined sets.

The highest predictive performance of RF is congruent with the existing literature in Learning Analytics, where RF consistently ranks among the best models \cite{hlosta_18,alcaraz_21}. To the best of our knowledge, our second best approach, however, Bayesian multilevel models with random slopes, is not mentioned in any LA studies, due to its nature as a statistical approach. Since most of the studies in LA are focused on predictions from objective data and BA-RS was the best algorithm in our case, it might be useful for other researchers to consider multilevel models in their work. After all, repeated measures are a common occurrence when assessing student data and multilevel models provide the means to reflect nested data structures. Nevertheless, these comparisons should be taken cautiously since the outcomes differ in both studies. 

Next, we compared the different types of predictors (RQ3). Predicting delay with objective predictors yielded better results than relying on the subjective variables alone. 
By mixing both types of predictors, we achieved a slightly higher predictive performance, meaning that some of the subjective predictors increase their relative predictive power when combining them with the objective predictors. These results are unsurprisingly in line with the previous study our data stems from \cite{Imhof2021}. The superiority of the objective predictors was evident in all the ML models regardless of the underlying algorithm, strengthening the finding of the previous study. Similarly, \cite{pca_pardo_2016} reported that more objective variables correlated with academic outcomes than subjective variables, and \cite{yu_20_fair} also found a higher accuracy of objective features to predict course and next semester outcomes. However, in contrast with the previous study, the advantage of combining the two sets of predictors was more apparent here since the exact ranking of the models was rather unclear before. 

As expected, the best results across all the performance metrics were achieved when the subjective and objective variables were combined. This is also in line with \cite{pca_pardo_2016,tempelaar_21}, despite the different levels of granularity between our study and theirs. While both of these studies focused on the final course outcome, we were interested in a much more fine-grained measure (the delay of individual assignments). In both cases, the authors reported an increase in explained variance when objective variables were added to the subjective variables. Our results demonstrate that this effect remains consistent across different ML algorithms when predicting delay. However, it remains unclear if combined data has a long-term predictive advantage. In \cite{yu_20_fair} for instance, the authors found that combined data only had higher predictive accuracy for short-term predictions. For long-term predictions (e.g., the outcome of the following semester), the authors determined that single objective predictors yielded better results. We suspect this may be due to fluctuations in students' learning-related dispositions, which cannot be fully captured by a single measurement.  
Despite their trait-like nature, it would thus be advisable to repeatedly assess subjective factors instead of exclusively relying on once-collected historical student data. This comes with its own risks, e.g., alienating students by a high frequency of reassessments.

Compared to \cite{pca_pardo_2016}, our results show that using ML algorithms might shuffle the order of importance of the subjective variables and the magnitude of their contribution. While APSS is the most predictive feature for subjective data, when combined with objective predictors, it is superseded by the GASE factor, which was the least important factor for subjective-only predictions. This makes it more challenging to determine which variables to favor in a more parsimonious model. In the work of \cite{pca_pardo_2016}, the order of importance remained consistent. We assume that this discrepancy could either be traced back to the covariance structure of the predictors and/or the complex nature of some of the implemented algorithms, particularly the underlying decision trees that empower both RF and GBM. In decision trees, more complex dependencies of the variables are considered when building the model. 

\subsection{Limitations} 
This study has two major limitations that need to be addressed. First off, the comparatively low predictive accuracy of the models (some even performing barely above chance) raises the question of why it is not as high as one might expect based on models in other studies. One issue is our rather diminutive sample size. This could be remedied in a number of ways, for example by collecting more data across multiple semesters with the same students. The data accumulated over time can then serve as historical data, serving as training data for future predictions. The other contributing factor is the unbalance of our data. Two thirds of the assignments being handed in on time is not necessarily a representative finding. Given the reported prevalence of procrastination in higher education \cite{Steel2007}, one could assume that there is more of an equilibrium in other populations. Another plausible possibility is a stricter enforcement of deadlines, resulting in an even higher proportion of timely submissions and thus more unbalanced data. 
The ML approaches involved in this type of prediction thus need to be stable enough to account for such differences in the studied populations.

The second issue concerns the models themselves. Models with objective predictors outperforming models with subjective predictors is not that surprising, considering that the outcome variable delay and our objective predictors are all state variables. 
When interpreting the results, it is also important to note that the two sets of variables are not directly comparable, mainly due to their different granularity (the subjective predictors being based on data assessed on the student level and the objective variables being located on the assignment level in two cases and the course level in one case). This is also reflected in the outcome variable, which shared its granularity only with a few predictors, which were also revealed to be the most important variables, both in this study and the original one \cite{Imhof2021}. 

\subsection{Implications \& future work} 
The results imply that future research in procrastination and other types of dilatory behavior should include ML algorithms such as RF, rather than relying on traditional statistical approaches alone. This holds true even for cases with low numbers of variables. In order to select the best variables to enhance predictions, we thus recommend not to refrain from implementing ML algorithms and evaluating data sets that include both subjective and objective predictors.
This is particularly crucial when applying such models in the field, e.g., for for real-time predictions, considering these two types of predictors may not be available simultaneously. 
While our data was all collected at the end of the semester, the same set of predictors could be split into early and late predictors if they were to be assessed in an LMS with real-time predictions in the future.

For instance, our subjective predictors are trait variables, meaning they are supposedly stable across certain periods, and could thus already be assessed at the start of the semester, making them viable as early predictors. Despite not being as strong or consistent as their objective counterparts in our models, the subjective predictors could still potentially have value since they may provide some clues long before the objective predictors become active. Even though a lot of the algorithms performed below chance when operating with subjective variables alone (meaning the performance was lower than simply assuming all assignments would be submitted on time), SVR-VS still managed to perform above chance.

As discussed in \cite{Imhof2021}, the objective predictors we used have the drawback of requiring information about students' activity across the semester, meaning predictions based on them cannot be made unless that data is accumulated (at least partially). This again highlights the utility of combining both types of predictors. Moreover, some of these variables change throughout the course (e.g., the number of clicks on an assignment), and so would the predictive effect of that variable when doing real-time predictions. 
From this perspective, the optimal points in time should be determined during the semester to achieve the best possible predictive performance that allows for instructional feedback and other interventions to have a positive impact on dilatory behavior and other performance metrics.

These results also imply that selecting a single algorithm and sticking with it for all models may not be the ideal path to take. Considering that a different approach was favored for each of the three categories, it may be advisable to employ multiple ML algorithms for real-time predictions, depending on the current stage of the semester (e.g., starting with SVR to work with early predictors, which can then be accompanied by a multilevel model for the late predictors or get replaced entirely by RF once all predictors become available).

Combined, our results show promise that these types of models could be used for real-time predictions of delay in the future, which would be a necessary step if the end goal is to provide timely interventions to reduce maladaptive forms of dilatory behavior (e.g., procrastination). However, the sample size is currently not large enough to provide accurate real-time predictions, meaning more data needs to be collected first, involving an expanded array of predictors. This could include task-specific factors (e.g., motivational aspects such as students' interest in a topic) or predictors related to students' time management skills. The latter could be achieved by analyzing behavioral patterns, which was successfully incorporated in a mixture models study by Park and colleagues \cite{park2018understanding}. Another promising avenue is to assess some of the learning-related factors objectively rather than subjectively, e.g., by replacing the self-directed learning questionnaire with clickstream-based indicators of self-regulated learning (see \cite{li2020using}).

\vspace{0.05in}
\section{Conclusion}
In conclusion, when applying ML prediction models in the field, comparisons between various algorithms are needed to determine which ones deliver the highest performance. The suitability of a given ML algorithm when predicting dilatory behavior depends on the type of predictor. While objective predictors work best with a statistical approach (Bayesian multilevel models), subjective predictors 
are better served with Gradient Boosting Machines. When both types of variables are combined, Random Forests were the preferable ML algorithm in our case. Future studies need to increase the predictive performance (e.g., by expanding the roster of predictors) to allow such models to be implemented in learning management systems, ultimately enabling real-time predictions during a semester. This can then serve as the basis for interventions aiming at reducing procrastination and promoting timely submissions.

\bibliographystyle{unsrtnat}
\bibliography{preprint_prediction}  

\begin{thebibliography}{59}
\providecommand{\natexlab}[1]{#1}
\providecommand{\url}[1]{\texttt{#1}}
\expandafter\ifx\csname urlstyle\endcsname\relax
  \providecommand{\doi}[1]{doi: #1}\else
  \providecommand{\doi}{doi: \begingroup \urlstyle{rm}\Url}\fi

\bibitem[Dumford and Miller(2018)]{Dumford2018}
Amber~D. Dumford and Angie~L. Miller.
\newblock {Online learning in higher education: exploring advantages and
  disadvantages for engagement}.
\newblock \emph{Journal of Computing in Higher Education}, 30\penalty0
  (3):\penalty0 452--465, dec 2018.
\newblock ISSN 18671233.
\newblock \doi{10.1007/s12528-018-9179-z}.
\newblock URL \url{https://doi.org/10.1007/s12528-018-9179-z}.

\bibitem[Steel(2007)]{Steel2007}
Piers Steel.
\newblock {The nature of procrastination: A meta-analytic and theoretical
  review of quintessential self-regulatory failure.}
\newblock \emph{Psychological Bulletin}, 133\penalty0 (1):\penalty0 65--94,
  2007.
\newblock ISSN 1939-1455.
\newblock \doi{10.1037/0033-2909.133.1.65}.
\newblock URL
  \url{http://doi.apa.org/getdoi.cfm?doi=10.1037/0033-2909.133.1.65}.

\bibitem[Tice and Baumeister(1997)]{Tice1997}
Dianne~M Tice and Roy~F. Baumeister.
\newblock {Longitudinal Study of Procrastination, Performance, Stress, and
  Health: The Costs and Benefits of Dawdling}.
\newblock \emph{Psychological Science}, 8\penalty0 (6):\penalty0 454--458, nov
  1997.
\newblock ISSN 0956-7976.
\newblock \doi{10.1111/j.1467-9280.1997.tb00460.x}.
\newblock URL
  \url{http://journals.sagepub.com/doi/10.1111/j.1467-9280.1997.tb00460.x}.

\bibitem[Doherty(2006)]{doherty2006}
William Doherty.
\newblock {An analysis of multiple factors affecting retention in Web-based
  community college courses}.
\newblock \emph{The Internet and Higher Education}, 9\penalty0 (4):\penalty0
  245--255, oct 2006.
\newblock ISSN 1096-7516.
\newblock \doi{10.1016/J.IHEDUC.2006.08.004}.

\bibitem[Ellis and Knaus(1977)]{ellis1977}
Albert Ellis and William~J. Knaus.
\newblock \emph{{Overcoming procrastination : or how to think and act
  rationally in spite of life's inevitable hassles}}.
\newblock Institute for Rational Living, 1977.
\newblock ISBN 0917476042.

\bibitem[Corkin et~al.(2011{\natexlab{a}})Corkin, Shirley, and
  Lindt]{corkin2011comparing}
Danya~M Corkin, L~Yu Shirley, and Suzanne~F Lindt.
\newblock Comparing active delay and procrastination from a self-regulated
  learning perspective.
\newblock \emph{Learning and Individual Differences}, 21\penalty0 (5):\penalty0
  602--606, 2011{\natexlab{a}}.

\bibitem[Pychyl(2008)]{Pychyl2008}
Timothy~A. Pychyl.
\newblock {Savouring the Flavours of Delay}.
\newblock \emph{English Studies in Canada}, 34\penalty0 (2-3):\penalty0 25, jun
  2008.
\newblock ISSN 0317-0802.

\bibitem[Hwang et~al.(2017)Hwang, Chuz, and Yin]{la_hwang_2017}
Gwo-Jen Hwang, Hui-Chun Chuz, and Chengjiu Yin.
\newblock {Objectives}, {Methodologies} and {Research} {Issues} of {Learning}
  {Analytics}.
\newblock \emph{in {Interactive} {Learning} {Environments}}, 25\penalty0
  (2):\penalty0 143 -- 146, March 2017.
\newblock ISSN 1744-5191.
\newblock \doi{https://doi.org/10.1080/10494820.2017.1287338}.

\bibitem[Azimi et~al.(2020)Azimi, Popa, and Cucić]{dt_azimi_2020}
Sepinoud Azimi, Carmen-Gabriela Popa, and Tatjana Cucić.
\newblock {Improving} {Students} {Performance} in {Small-Scale} {Online}
  {Courses} - {A} {Machine} {Learning-Based} {Intervention}.
\newblock \emph{in {International} {Journal} of {Learning} {Analytics} and
  {Artificial} {Intelligence} for {Education} {(iJAI)}}, 2\penalty0
  (2):\penalty0 80 -- 95, November 2020.
\newblock \doi{10.3991/ijai.v2i2.19371}.

\bibitem[Pardo et~al.(2017)Pardo, Han, and Ellis]{pca_pardo_2016}
Abelardo Pardo, Feifei Han, and Robert~A. Ellis.
\newblock {Combining} {University} {Student} {Self-Regulated} {Learning}
  {Indicators} and {Engagement} with {Online} {Learning} {Events} to {Predict}
  {Academic} {Performance}.
\newblock \emph{in {IEEE} {Transactions} on {Learning} {Technologies}},
  10\penalty0 (1):\penalty0 82 -- 92, March 2017.
\newblock ISSN 1939-1382.
\newblock \doi{10.1109/TLT.2016.2639508}.

\bibitem[Conijn et~al.(2017)Conijn, Snijders, Kleingeld, and
  Matzat]{olsr_conijn_2016}
Rianne Conijn, Chris Snijders, Ad~Kleingeld, and Uwe Matzat.
\newblock {Predicting} {Student} {Performance} from {LMS} {Data}: {A}
  {Comparison} of 17 {Blended} {Courses} {Using} {Moodle} {LMS}.
\newblock \emph{in {IEEE} {Transactions} on {Learning} {Technologies}},
  10\penalty0 (1):\penalty0 17 -- 29, March 2017.
\newblock ISSN 1939-1382.
\newblock \doi{10.1109/TLT.2016.2616312}.

\bibitem[Akram et~al.(2019)Akram, Fu, Li, Javed, Lin, Jiang, and
  Tang]{Akram2019}
Aftab Akram, Chengzhou Fu, Yuyao Li, Muhammad~Yaqoob Javed, Ronghua Lin,
  Yuncheng Jiang, and Yong Tang.
\newblock {Predicting Students' Academic Procrastination in Blended Learning
  Course Using Homework Submission Data}.
\newblock \emph{IEEE Access}, 7:\penalty0 102487--102498, 2019.
\newblock ISSN 2169-3536.
\newblock \doi{10.1109/access.2019.2930867}.

\bibitem[Cerezo et~al.(2017)Cerezo, Esteban, S{\'{a}}nchez-Santill{\'{a}}n, and
  N{\'{u}}{\~{n}}ez]{Cerezo2017}
Rebeca Cerezo, Mar{\'{i}}a Esteban, Miguel S{\'{a}}nchez-Santill{\'{a}}n, and
  Jos{\'{e}}~C. N{\'{u}}{\~{n}}ez.
\newblock {Procrastinating behavior in computer-based learning environments to
  predict performance: A case study in Moodle}.
\newblock \emph{Frontiers in Psychology}, 8\penalty0 (AUG):\penalty0 1--11,
  2017.
\newblock ISSN 16641078.
\newblock \doi{10.3389/fpsyg.2017.01403}.

\bibitem[Imhof et~al.(2021)Imhof, Bergamin, and McGarrity]{Imhof2021}
Christof Imhof, Per Bergamin, and Stéphanie McGarrity.
\newblock {Prediction of dilatory behaviour in online assignments}.
\newblock \emph{Learning and Individual Differences}, 88, May 2021.
\newblock \doi{10.1016/j.lindif.2021.102014}.

\bibitem[Corkin et~al.(2011{\natexlab{b}})Corkin, Yu, and Lindt]{Corkin2011}
Danya~M. Corkin, Shirley~L. Yu, and Suzanne~F. Lindt.
\newblock {Comparing active delay and procrastination from a self-regulated
  learning perspective}.
\newblock \emph{Learning and Individual Differences}, 21\penalty0 (5):\penalty0
  602--606, 2011{\natexlab{b}}.
\newblock ISSN 10416080.
\newblock \doi{10.1016/j.lindif.2011.07.005}.
\newblock URL \url{http://dx.doi.org/10.1016/j.lindif.2011.07.005}.

\bibitem[Bandura(1986)]{Bandura1986}
Albert Bandura.
\newblock \emph{{Social foundations of thought and action: A social cognitive
  theory.}}
\newblock Prentice-Hall series in social learning theory. Prentice-Hall, Inc,
  Englewood Cliffs, NJ, US, 1986.
\newblock ISBN 0-13-815614-X (Hardcover).

\bibitem[W{\"{a}}schle et~al.(2014)W{\"{a}}schle, Allgaier, Lachner, Fink, and
  N{\"{u}}ckles]{Waschle2014a}
Kristin W{\"{a}}schle, Anne Allgaier, Andreas Lachner, Siegfried Fink, and
  Matthias N{\"{u}}ckles.
\newblock {Procrastination and self-efficacy: Tracing vicious and virtuous
  circles in self-regulated learning}.
\newblock \emph{Learning and Instruction}, 29:\penalty0 103--114, 2014.
\newblock ISSN 09594752.
\newblock \doi{10.1016/j.learninstruc.2013.09.005}.
\newblock URL \url{http://dx.doi.org/10.1016/j.learninstruc.2013.09.005}.

\bibitem[Chu and Choi(2005)]{Chu2005}
Angela Hsin~Chun Chu and Jin~Nam Choi.
\newblock {Rethinking procrastination: positive effects of "active"
  procrastination behavior on attitudes and performance.}
\newblock \emph{The Journal of Social Psychology}, 145\penalty0 (3):\penalty0
  245--64, 2005.
\newblock ISSN 0022-4545.
\newblock \doi{10.3200/SOCP.145.3.245-264}.

\bibitem[Bolhuis(1996)]{Bolhuis1996}
Sanneke Bolhuis.
\newblock {Towards Active and Selfdirected Learning. Preparing for Lifelong
  Learning, with Reference to Dutch Secondary Education.}, apr 1996.

\bibitem[Knowles(1975)]{Knowles1975}
Malcolm~Shepherd Knowles.
\newblock \emph{{Self-directed learning : a guide for learners and teachers}}.
\newblock Chicago : Association Press, 1975.

\bibitem[Garrison(1997)]{Garrison1997}
D.~R. Garrison.
\newblock {Self-Directed Learning: Toward a Comprehensive Model}.
\newblock \emph{Adult Education Quarterly}, 48\penalty0 (1):\penalty0 18--33,
  nov 1997.
\newblock ISSN 0741-7136.
\newblock \doi{10.1177/074171369704800103}.
\newblock URL \url{http://journals.sagepub.com/doi/10.1177/074171369704800103}.

\bibitem[Saeid and Eslaminejad(2017)]{Saeid2017}
Nasim Saeid and Tahere Eslaminejad.
\newblock {Relationship between Student's Self-Directed-Learning Readiness and
  Academic Self-Efficacy and Achievement Motivation in Students}.
\newblock \emph{International Education Studies}, 10\penalty0 (1), 2017.
\newblock ISSN 1913-9039.
\newblock \doi{10.5539/ies.v10n1p225}.

\bibitem[Schommer-Aikins and Easter(2018)]{Schommer2018cognitive}
Marlene Schommer-Aikins and Marilyn Easter.
\newblock {Cognitive flexibility, procrastination, and need for closure linked
  to online self-directed learning among students taking online courses}.
\newblock \emph{Journal of Business and Educational Leadership}, 8\penalty0
  (1):\penalty0 112--121, 2018.

\bibitem[Cirigliano et~al.(2020)Cirigliano, Guthrie, and Pusic]{Cirigliano2020}
Matthew~M. Cirigliano, Charles~D. Guthrie, and Martin~V. Pusic.
\newblock {Click-level Learning Analytics in an Online Medical Education
  Learning Platform}.
\newblock \emph{Teaching and Learning in Medicine}, 32\penalty0 (4):\penalty0
  410--421, aug 2020.
\newblock ISSN 1040-1334.
\newblock \doi{10.1080/10401334.2020.1754216}.
\newblock URL
  \url{https://www.tandfonline.com/doi/full/10.1080/10401334.2020.1754216}.

\bibitem[Knight et~al.(2017)Knight, {Friend Wise}, and Chen]{Knight2017}
Simon Knight, Alyssa {Friend Wise}, and Bodong Chen.
\newblock {Time for Change: Why Learning Analytics Needs Temporal Analysis}.
\newblock \emph{Journal of Learning Analytics}, 4\penalty0 (3):\penalty0 7--17,
  dec 2017.
\newblock \doi{10.18608/jla.2017.43.2}.
\newblock URL \url{http://dx.doi.org/10.18608/}.

\bibitem[You(2015)]{You2015}
Ji~Won You.
\newblock {Examining the Effect of Academic Procrastination on Achievement
  Using LMS in e-Learning}.
\newblock \emph{Educational Technology {\&} Society}, 18\penalty0 (3):\penalty0
  64--74, 2015.
\newblock ISSN 1436-4522.

\bibitem[Levy and Ramim(2012)]{Levy2012}
Yair Levy and Michelle~M. Ramim.
\newblock {A Study of Online Exams Procrastination Using Data Analytics
  Techniques}.
\newblock \emph{Interdisciplinary Journal of e-Skills and Lifelong Learning},
  8:\penalty0 097--113, 2012.
\newblock ISSN 2375-2084.
\newblock \doi{10.28945/1730}.

\bibitem[Gareau et~al.(2019)Gareau, Chamandy, Kljajic, and
  Gaudreau]{Gareau2019}
Alexandre Gareau, M{\'e}lodie Chamandy, Kristina Kljajic, and Patrick Gaudreau.
\newblock The detrimental effect of academic procrastination on subsequent
  grades: the mediating role of coping over and above past achievement and
  working memory capacity.
\newblock \emph{Anxiety, Stress, \& Coping}, 32\penalty0 (2):\penalty0
  141--154, 2019.

\bibitem[Yamada et~al.(2016)Yamada, Goda, Matsuda, Saito, Kato, and
  Miyagawa]{Yamada2016}
Masanori Yamada, Yoshiko Goda, Takeshi Matsuda, Yutaka Saito, Hiroshi Kato, and
  Hiroyuki Miyagawa.
\newblock {How does self-regulated learning relate to active procrastination
  and other learning behaviors?}
\newblock \emph{Journal of Computing in Higher Education}, 28\penalty0
  (3):\penalty0 326--343, 2016.
\newblock ISSN 18671233.
\newblock \doi{10.1007/s12528-016-9118-9}.

\bibitem[{del Puerto Paule-Ruiz} et~al.(2015){del Puerto Paule-Ruiz},
  Riestra-González, Sánchez-Santillán, and
  Pérez-Pérez]{DelPuertoPaule-Ruiz2015}
María {del Puerto Paule-Ruiz}, Moises Riestra-González, Miguel
  Sánchez-Santillán, and Juan~Ramón Pérez-Pérez.
\newblock {The procrastination related indicators in e-learning platforms}.
\newblock \emph{Journal of Universal Computer Science}, 21\penalty0
  (1):\penalty0 7--22, 2015.
\newblock ISSN 09486968.

\bibitem[Hooshyar et~al.(2020)Hooshyar, Pedaste, and Yang]{Hooshyar2020mining}
Danial Hooshyar, Margus Pedaste, and Yeongwook Yang.
\newblock Mining educational data to predict students’ performance through
  procrastination behavior.
\newblock \emph{Entropy}, 22\penalty0 (1):\penalty0 12, 2020.

\bibitem[Abidi et~al.(2020)Abidi, Zhang, Haidery, Rizvi, Riaz, Ding, and
  Kwon]{abidi2020educational}
Syed Muhammad~Raza Abidi, Wu~Zhang, Saqib~Ali Haidery, Sanam~Shahla Rizvi,
  Rabia Riaz, Hu~Ding, and Se~Jin Kwon.
\newblock Educational sustainability through big data assimilation to quantify
  academic procrastination using ensemble classifiers.
\newblock \emph{Sustainability}, 12\penalty0 (15):\penalty0 6074, 2020.

\bibitem[Yang et~al.(2020)Yang, Hooshyar, Pedaste, Wang, Huang, and
  Lim]{Yang2020a}
Yeongwook Yang, Danial Hooshyar, Margus Pedaste, Minhong Wang, Yueh~Min Huang,
  and Heuiseok Lim.
\newblock {Prediction of students' procrastination behaviour through their
  submission behavioural pattern in online learning}.
\newblock \emph{Journal of Ambient Intelligence and Humanized Computing}, 2020.
\newblock ISSN 18685145.
\newblock \doi{10.1007/s12652-020-02041-8}.
\newblock URL \url{https://doi.org/10.1007/s12652-020-02041-8}.

\bibitem[Nielsen et~al.(2018)Nielsen, Dammeyer, Vang, and
  Makransky]{Nielsen2018}
T.~Nielsen, J.~Dammeyer, M.~L. Vang, and G.~Makransky.
\newblock {Gender Fairness in Self-Efficacy? A Rasch-Based Validity Study of
  the General Academic Self-Efficacy Scale (GASE)}.
\newblock \emph{Scandinavian Journal of Educational Research}, 62\penalty0
  (5):\penalty0 664--681, 2018.
\newblock ISSN 14701170.
\newblock \doi{10.1080/00313831.2017.1306796}.

\bibitem[Lounsbury et~al.(2009)Lounsbury, Levy, Park, Gibson, and
  Smith]{Lounsbury2009}
John~W. Lounsbury, Jacob~J. Levy, Soo~Hee Park, Lucy~W. Gibson, and Ryan Smith.
\newblock {An investigation of the construct validity of the personality trait
  of self-directed learning}.
\newblock \emph{Learning and Individual Differences}, 19\penalty0 (4):\penalty0
  411--418, 2009.
\newblock ISSN 10416080.
\newblock \doi{10.1016/j.lindif.2009.03.001}.
\newblock URL \url{http://dx.doi.org/10.1016/j.lindif.2009.03.001}.

\bibitem[McCloskey(2012)]{McCloskey2011}
Justin~D. McCloskey.
\newblock \emph{{Finally, My Thesis On Academic Procrastination}}.
\newblock PhD thesis, 2012.

\bibitem[Choi and Moran(2009)]{Choi2009}
Jin~Nam Choi and Sarah~V Moran.
\newblock {Why not procrastinate? Development and validation of a new active
  procrastination scale.}
\newblock \emph{The Journal of social psychology}, 149\penalty0 (2):\penalty0
  195--211, 2009.
\newblock ISSN 0022-4545.
\newblock \doi{10.3200/SOCP.149.2.195-212}.

\bibitem[Wright et~al.(2017)Wright, Dankowski, and
  Ziegler]{maxstat_wright_2017}
Marvin~N Wright, Theresa Dankowski, and Andreas Ziegler.
\newblock Unbiased split variable selection for random survival forests using
  maximally selected rank statistics.
\newblock \emph{Statistics in Medicine}, 36\penalty0 (8):\penalty0 1272--1284,
  2017.

\bibitem[Friedman(2001)]{Friedman:2001:GBM}
Jerome~H Friedman.
\newblock Greedy function approximation: a gradient boosting machine.
\newblock \emph{Annals of statistics}, pages 1189--1232, 2001.

\bibitem[Friedman(2002)]{Friedman:2002:StochasticGBM}
Jerome~H Friedman.
\newblock Stochastic gradient boosting.
\newblock \emph{Computational statistics \& data analysis}, 38\penalty0
  (4):\penalty0 367--378, 2002.

\bibitem[Hlosta et~al.(2021)Hlosta, Herodotou, Bayer, and
  Fernandez]{Hlosta:2021:ImpactGBM}
Martin Hlosta, Christothea Herodotou, Vaclav Bayer, and Miriam Fernandez.
\newblock Impact of predictive learning analytics on course awarding gap of
  disadvantaged students in stem.
\newblock In \emph{International Conference on Artificial Intelligence in
  Education}, pages 190--195. Springer, 2021.

\bibitem[Ruip{\'e}rez-Valiente et~al.(2017)Ruip{\'e}rez-Valiente, Cobos,
  Mu{\~n}oz-Merino, Andujar, and Kloos]{Ruiperez:2017:earlyMOOC}
Jos{\'e}~A Ruip{\'e}rez-Valiente, Ruth Cobos, Pedro~J Mu{\~n}oz-Merino,
  {\'A}lvaro Andujar, and Carlos~Delgado Kloos.
\newblock Early prediction and variable importance of certificate
  accomplishment in a mooc.
\newblock In \emph{European Conference on Massive Open Online Courses}, pages
  263--272. Springer, 2017.

\bibitem[2015(2015)]{KDDcup:2015}
KDD~Cup 2015.
\newblock Kdd cup 2015: Predicting dropouts in mooc, 2015.

\bibitem[Wright and Ziegleri(2017)]{wright2017}
M.~N. Wright and A.~Ziegleri.
\newblock ranger: A fast implementation of random forests for high dimensional
  data in c++ and r.
\newblock \emph{Journal of Statistical Software}, 77\penalty0 (1):\penalty0 1
  -- 17, 2017.

\bibitem[Bürkner(2017)]{Buerkner2017brms}
Paul-Christian Bürkner.
\newblock {brms}: An {R} package for {Bayesian} multilevel models using {Stan}.
\newblock \emph{Journal of Statistical Software}, 80\penalty0 (1):\penalty0
  1--28, 2017.
\newblock \doi{10.18637/jss.v080.i01}.

\bibitem[Pedregosa et~al.(2011)Pedregosa, Varoquaux, Gramfort, Michel, Thirion,
  Grisel, Blondel, Prettenhofer, Weiss, Dubourg, et~al.]{pedregosa2011scikit}
Fabian Pedregosa, Ga{\"e}l Varoquaux, Alexandre Gramfort, Vincent Michel,
  Bertrand Thirion, Olivier Grisel, Mathieu Blondel, Peter Prettenhofer, Ron
  Weiss, Vincent Dubourg, et~al.
\newblock Scikit-learn: Machine learning in python.
\newblock \emph{the Journal of machine Learning research}, 12:\penalty0
  2825--2830, 2011.

\bibitem[Parrella(2007)]{parella2007}
Francesco Parrella.
\newblock \emph{Online support vector regression [master's thesis]}.
\newblock PhD thesis, University of Genoa, Italy, 2007.

\bibitem[Comsa et~al.(2021)Comsa, Molnar, Tal, Bergamin, Muntean, Muntean, and
  Trestian]{comsa_education_2021}
I.-S. Comsa, A.~Molnar, I.~Tal, P.~Bergamin, C.~Hava Muntean, G.-M. Muntean,
  and R.~Trestian.
\newblock A {Machine} {Learning} {Resource} {Allocation} {Solution} to
  {Improve} {Video} {Quality} in {Remote} {Education}.
\newblock \emph{in {IEEE} {Transactions} on {Broadcasting}}, pages 1 -- 21,
  April 2021.
\newblock ISSN 1557-9611.
\newblock \doi{10.1109/TBC.2021.3068872}.

\bibitem[Comsa(2014)]{comsa_thesis_2014}
Ioan-Sorin Comsa.
\newblock \emph{{Sustainable} {Scheduling} {Policies} for {Radio} {Access}
  {Networks} {Based} on {LTE} {Technology}}.
\newblock University of Bedfordshire, U.K., 2014.
\newblock ISBN PhD Thesis.

\bibitem[Rousseeuw(1987)]{si_index}
Peter~J. Rousseeuw.
\newblock {Silhouette}s: {A} graphical aid to the interpretation and validation
  of cluster analysis.
\newblock \emph{Journal of Computational and Applied Mathematics}, 20:\penalty0
  53 -- 65, 1987.

\bibitem[Chicco and Jurman(2020)]{chicco2020advantages}
Davide Chicco and Giuseppe Jurman.
\newblock The advantages of the matthews correlation coefficient (mcc) over f1
  score and accuracy in binary classification evaluation.
\newblock \emph{BMC genomics}, 21\penalty0 (1):\penalty0 1--13, 2020.

\bibitem[Chicco(2017)]{chicco2017ten}
Davide Chicco.
\newblock Ten quick tips for machine learning in computational biology.
\newblock \emph{BioData mining}, 10\penalty0 (1):\penalty0 1--17, 2017.

\bibitem[T.~Daniya and Kumar(2020)]{gini_index_2021}
M.~Geetha T.~Daniya and K.~Suresh Kumar.
\newblock {Classification} and {Regression} {Trees} with {Gini} {Index}.
\newblock \emph{in {Advances} in {Mathematics}: {Scientific} {Journal}},
  9\penalty0 (10):\penalty0 8237 -- 8247, 2020.
\newblock ISSN 1857-8438.
\newblock \doi{https://doi.org/10.37418/amsj.9.10.53}.

\bibitem[Hlosta et~al.(2018)Hlosta, Zdrahal, and Zendulka]{hlosta_18}
Martin Hlosta, Zdenek Zdrahal, and Jaroslav Zendulka.
\newblock Are we meeting a deadline? classification goal achievement in time in
  the presence of imbalanced data.
\newblock \emph{Knowledge-Based Systems}, 160:\penalty0 278--295, 2018.

\bibitem[Alcaraz et~al.(2021)Alcaraz, Martinez-Rodrigo, Zangroniz, and
  Rieta]{alcaraz_21}
Raul Alcaraz, Arturo Martinez-Rodrigo, Roberto Zangroniz, and Jose~Joaquin
  Rieta.
\newblock Early prediction of students at risk of failing a face-to-face course
  in power electronic systems.
\newblock \emph{IEEE Transactions on Learning Technologies}, 2021.

\bibitem[Yu et~al.(2020)Yu, Li, Fischer, Doroudi, and Xu]{yu_20_fair}
Renzhe Yu, Qiujie Li, Christian Fischer, Shayan Doroudi, and Di~Xu.
\newblock Towards accurate and fair prediction of college success: Evaluating
  different sources of student data.
\newblock \emph{International Educational Data Mining Society}, 2020.

\bibitem[Tempelaar et~al.(2021)Tempelaar, Rienties, and Nguyen]{tempelaar_21}
Dirk Tempelaar, Bart Rienties, and Quan Nguyen.
\newblock The contribution of dispositional learning analytics to precision
  education.
\newblock \emph{Journal of Educational Technology and Society}, 24\penalty0
  (1):\penalty0 109--122, January 2021.

\bibitem[Park et~al.(2018)Park, Yu, Rodriguez, Baker, Smyth, and
  Warschauer]{park2018understanding}
Jihyun Park, Renzhe Yu, Fernando Rodriguez, Rachel Baker, Padhraic Smyth, and
  Mark Warschauer.
\newblock Understanding student procrastination via mixture models.
\newblock \emph{International Educational Data Mining Society}, 2018.

\bibitem[Li et~al.(2020)Li, Baker, and Warschauer]{li2020using}
Qiujie Li, Rachel Baker, and Mark Warschauer.
\newblock Using clickstream data to measure, understand, and support
  self-regulated learning in online courses.
\newblock \emph{The Internet and Higher Education}, 45:\penalty0 100727, 2020.

\end{thebibliography}






\vspace{0.5in}
\noindent\begin{minipage}{0.3\textwidth}
\includegraphics[width=1.0in]{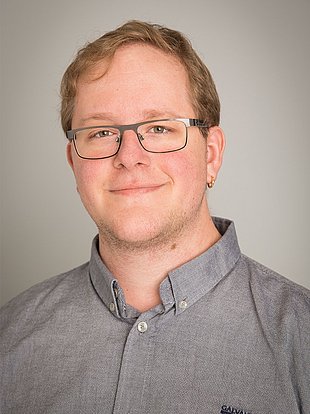}
\end{minipage}
\hfill
\begin{minipage}{0.6\textwidth}
{Christof Imhof} has been employed at the Institute for Research in Open, Distance, and e-Learning (IFeL) since 2016 and is currently a doctoral student at the University of Bern. His research focus lies primarily on procrastination and other types of dilatory behavior in the context of adaptive learning, which also serves as the topic of his doctoral thesis. Other research interests include the detection of emotions with objective measures such as eye-tracking combined with emotional word lists.
\end{minipage}

\vspace{0.5in}
\noindent\begin{minipage}{0.3\textwidth}
\includegraphics[width=1.0in]{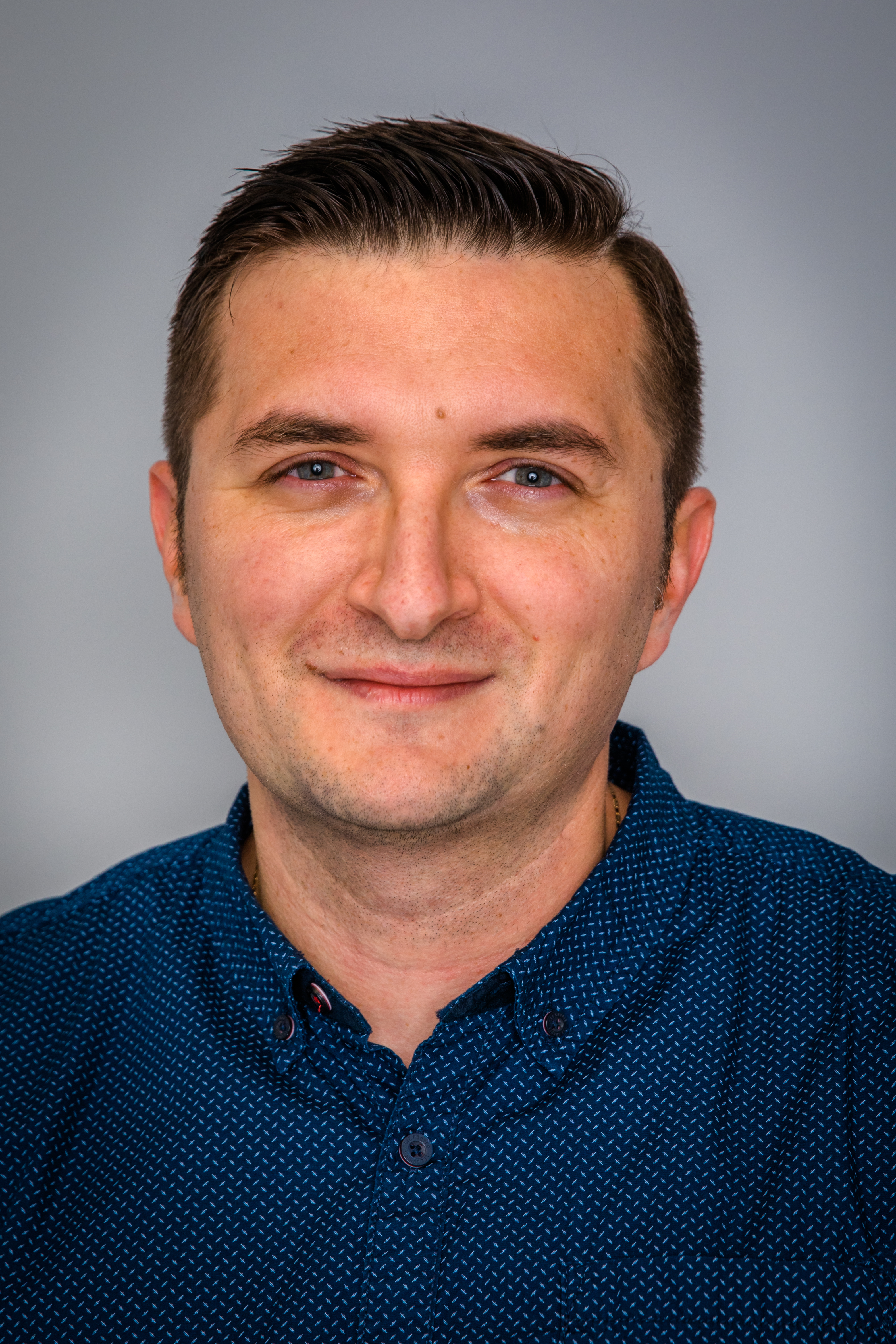}
\end{minipage}
\hfill
\begin{minipage}{0.6\textwidth}
{Ioan-Sorin Comsa} is a Data Scientist with the Institute for Open, Distance and eLearning (IFeL). He received a joint PhD from the University of Bedfordshire, U.K. and the University of Applied Sciences of Western Switzerland, in 2015. He worked as a Senior Research Engineer with CEA-LETI, France. Since 2017, he has been a Research Assistant with Brunel University London being actively involved in EU NEWTON project. His research interests include personalised and adaptive learning, machine learning and wireless communications.
\end{minipage}

\vspace{0.5in}
\noindent\begin{minipage}{0.3\textwidth}
\includegraphics[width=1.0in]{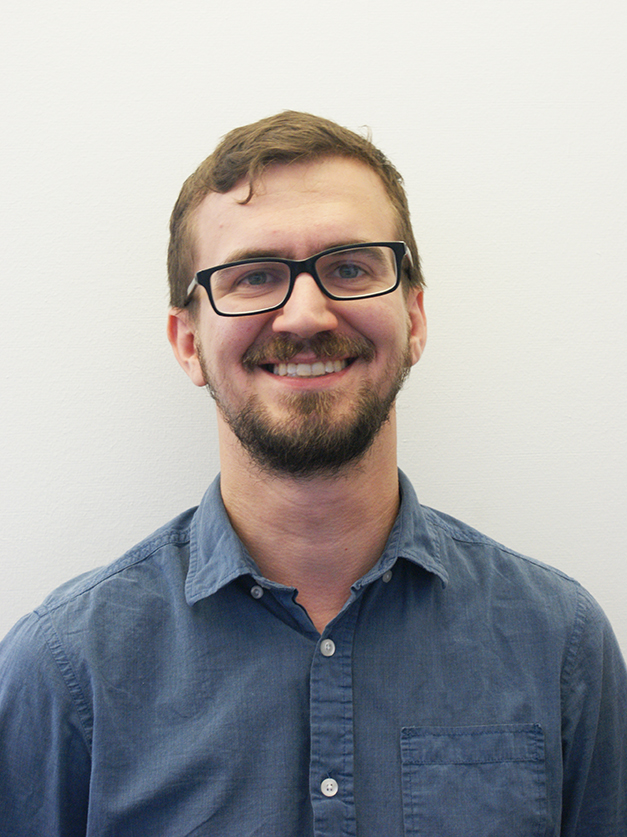}
\end{minipage}
\hfill
\begin{minipage}{0.6\textwidth}
{Martin Hlosta} is a research fellow at the Institut for Distance Learning and eLearning Research (IFeL) working on projects for adaptive learning in education. Before, he led OUAnalyse at The Open University, a project improving student retention via machine learning, selected by UNESCO among four of the best projects using AI in Education in 2020. His research mainly focuses on Predictive Learning Analytics and its impact, in particular to target existing educational inequalities. 
\end{minipage}

\vspace{0.5in}
\noindent\begin{minipage}{0.3\textwidth}
\includegraphics[width=1.0in]{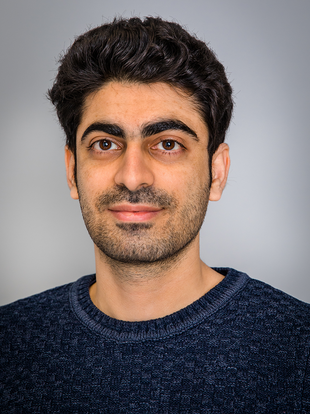}
\end{minipage}
\hfill
\begin{minipage}{0.6\textwidth}
{Behnam Parsaeifard} has joined the Institute for Open-, Distance- and eLearning (IFeL) as a researcher in November 2021. Prior to that, he received his PhD in computational physics from the University of Basel and was a research intern at the Visual Intelligence for Transportation (VITA) lab of EPFL. He is interested in applying data mining and machine learning techniques to improve adaptive learning and learning analytics.
\end{minipage}

\vspace{0.5in}
\noindent\begin{minipage}{0.3\textwidth}
\includegraphics[width=1.0in]{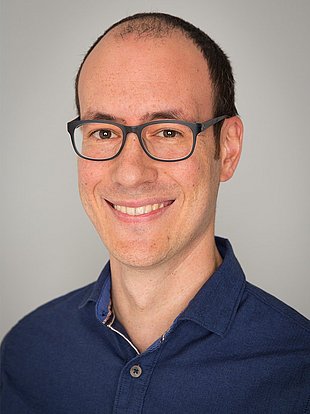}
\end{minipage}
\hfill
\begin{minipage}{0.6\textwidth}
{Ivan Moser} is a Researcher and Lecturer at the Institute for Research in Open, Distance and eLearning (IFeL). Previously, he studied Psychology at the University of Bern, where he earned his PhD at the Institute of Cognitive Psychology. In the course of his research career, Ivan Moser developed an increasing fascination for the numerous new possibilities that virtual reality (VR) opens as a learning technology. As leader of several VR-related research projects at FFHS, he investigates the potential and limits of virtual learning environments.
\end{minipage}

\vspace{0.5in}
\noindent\begin{minipage}{0.3\textwidth}
\includegraphics[width=1.0in]{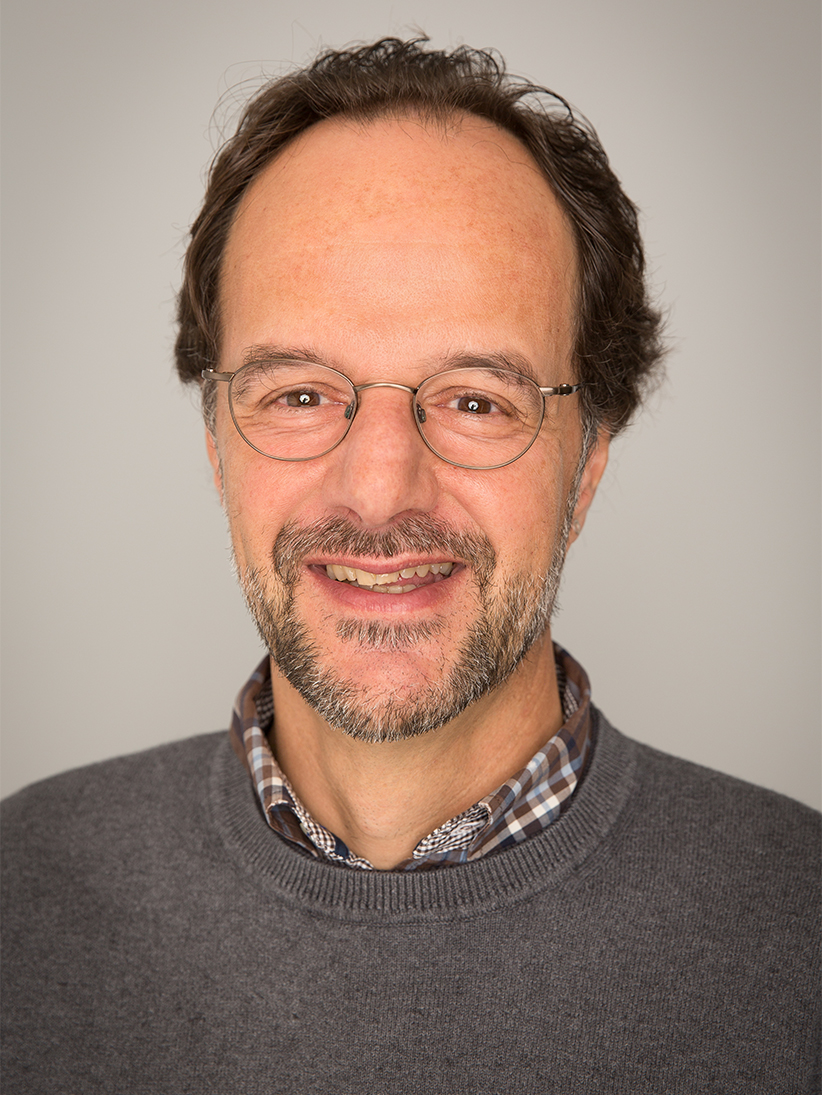}
\end{minipage}
\hfill
\begin{minipage}{0.6\textwidth}
{Per Bergamin} is Professor for Didactics in Distance Education and E-Learning at the Swiss Distance University of Applied Sciences (FFHS). Since 2006 he acts as the Director of the Institute for Research in Open-, Distance- and eLearning (IFeL) and from 2016 on he holds also the UNESCO Chair on personalised and adaptive Distance Education. His research activities focus on self-regulated and technology-based personalized and adaptive learning. Central aspects are instructional design, usability and application implementation.
\end{minipage}

\end{document}